%% file: main.tex
\documentclass{article}

\usepackage[final]{neurips_2024}
\usepackage{float}

\usepackage[utf8]{inputenc} % allow utf-8 input
\usepackage[T1]{fontenc}    % use 8-bit T1 fonts
\usepackage[pagebackref=false,breaklinks,colorlinks,citecolor=gray]{hyperref}
\usepackage{url}            % simple URL typesetting
\usepackage{booktabs}       % professional-quality tables
\usepackage{amsfonts}       % blackboard math symbols
\usepackage{nicefrac}       % compact symbols for 1/2, etc.
\usepackage{microtype}      % microtypography
\usepackage{amsmath}
\usepackage{graphicx}
\usepackage{subcaption}
\usepackage{enumitem}
\usepackage{ragged2e}
\usepackage[table,xcdraw]{xcolor}
\usepackage[utf8]{inputenc}
\usepackage{tcolorbox}
\usepackage{enumitem}
\usepackage{lipsum} % For placeholder text, can be removed
\usepackage{bookmark}

\tcbset{
  colback=gray!5,
  colframe=black!50,
  fonttitle=\bfseries,
  coltitle=black,
  boxrule=0.5pt,
  arc=2mm, % <--- 모서리 둥근 정도 (0mm = 각짐, 2~3mm = 적당히 둥글게)
  left=5pt,
  right=5pt,
  top=5pt,
  bottom=5pt,
  boxsep=5pt
}

\usepackage[utf8]{inputenc}
\usepackage{tcolorbox}
\tcbuselibrary{breakable}
\usepackage{enumitem}
\usepackage{geometry}
\usepackage{lipsum}
\definecolor{pastelblue}{HTML}{D8E6FC}
\definecolor{pastelblue2}{HTML}{F8FBFF}

\title{Cross-Frame Representation Alignment for Fine-Tuning Video Diffusion Models}

\author{%
  Sungwon Hwang\thanks{Equal Contribution}\quad Hyojin Jang$^{*}$\quad Kinam Kim\quad Minho Park\quad Jaegul Choo \\
  KAIST AI \\
  \url{https://crepavideo.github.io} \\
}

\begin{document}

\maketitle

\begin{figure*}[h]
\centering
\vspace{-0.75cm}
\includegraphics[width=0.95\textwidth]{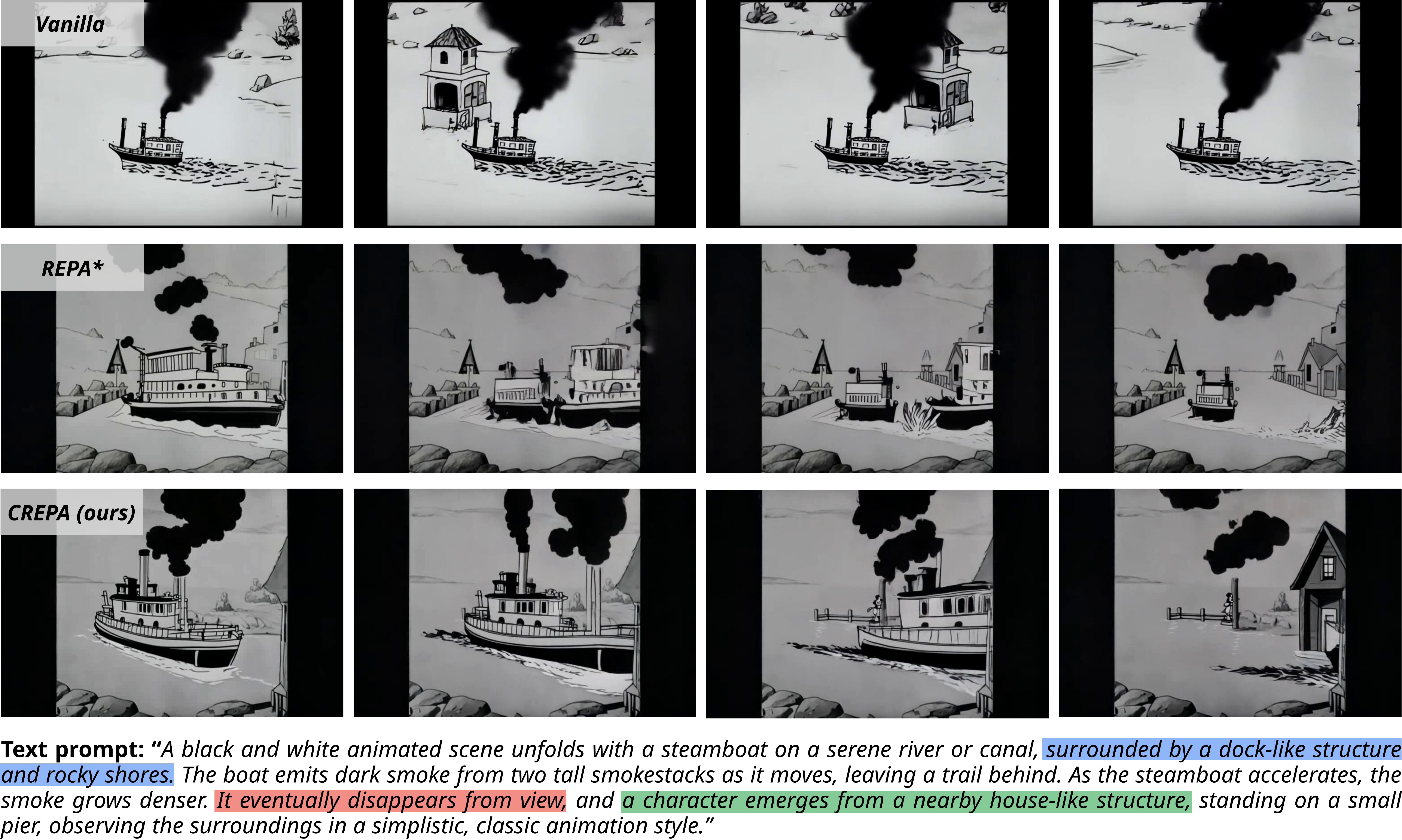}
\captionof{figure}{\textbf{Videos generated by CogVideoX-5B~\cite{yang2024cogvideox} fine-tuned on the Disney~\cite{wildheart2024disney} dataset}. Each model is fine-tuned with: no regularization (Vanilla), REPA* (our implementation of REPA~\cite{yu2024representation} to video diffusion models), and CREPA \textit{(ours)}. Our model yields beter text reflectivity and semantic consistency across frames compared to the baselines.}
% \captionof{figure}{\textbf{Videos generated by CogVideoX-5B~\cite{yang2024cogvideox} fine-tuned on the Disney~\cite{wildheart2024disney} dataset}. Each model is fine-tuned with: no regularization (Vanilla), REPA* (our implementation of REPA~\cite{yu2024representation} to video diffusion models), and CREPA \textit{(ours)}. CREPA successfully induces the model to generate semantically coherent object structures across frames compared to REPA*, while showing better text reflectivity, which indicates better convergence, compared to Vanilla.}
\label{fig:teaser}
\end{figure*}

\input{sections/0_abstract}
\input{sections/1_intro}

\input{sections/2_related_works}

\input{sections/3_prelim}
\input{sections/4_methods}
\input{sections/5_experiments}
\input{sections/6_conclusion_and_future_works}

{\small
\bibliographystyle{abbrv}
\bibliography{bibliography}
}

\input{sections/7_appendix}

\end{document}

%% file: sections/0_abstract.tex
\begin{abstract}

Fine-tuning Video Diffusion Models (VDMs) at the user level to generate videos that reflect specific attributes of training data presents notable challenges, yet remains underexplored despite its practical importance. Meanwhile, recent work such as Representation Alignment (REPA) has shown promise in improving the convergence and quality of DiT-based image diffusion models by aligning, or assimilating, its internal hidden states with external pretrained visual features, suggesting its potential for VDM fine-tuning. In this work, we first propose a straightforward adaptation of REPA for VDMs and empirically show that, while effective for convergence, it is suboptimal in preserving semantic consistency across frames. To address this limitation, we introduce \textit{Cross-frame Representation Alignment} (CREPA), a novel regularization technique that aligns hidden states of a frame with external features from neighboring frames. Empirical evaluations on large-scale VDMs, including CogVideoX-5B and Hunyuan Video, demonstrate that CREPA improves both visual fidelity and cross-frame semantic coherence when fine-tuned with parameter-efficient methods such as LoRA. We further validate CREPA across diverse datasets with varying attributes, confirming its broad applicability. 
% Project page: \url{https://crepa.github.io}.

% Fine-tuning Video Diffusion Models (VDMs) at user level to generate videos with specific attributes of training data poses significant challenges. However, the problem remains underexplored despite its practical significance. Meanwhile, recently proposed distillation method such as Representation Alignment (REPA) have improved the convergence speed and quality for training image diffusion models by aligning their internal representations with pretrained, high quality external visual features, suggesting its potential applicability to fine-tuning VDMs. However, we learned that REPA is suboptimal for VDMs as it does not guarantee smooth and consistent semantics between adjacent frames during the alignment. To address this issue, we propose \textit{Cross-frame Representation Alignment} (CREPA), a novel regularization method for VDMs that explicitly incorporates temporal context of data by additionally regularizing hidden states of a frame toward external pretrained features of adjacent frames. Empirical evaluations show that CREPA enhances visual fidelity and consistent semantics when fine-tuning large-scale video diffusion models such as CogVideoX-5B and Hunyuan Video using parameter-efficient approaches such as LoRA. We further validate our method by fine-tuning and evaluating on diverse datasets containing various visual attributes, demonstrating its effectiveness.
\end{abstract}

% \begin{abstract}
% Fine-tuning Video Diffusion Models (VDMs) to generate videos with specific attributes of training data poses significant challenges due to the limited training steps and computational constraints at the user level. While recent advancements such as Representation Alignment (REPA) have improved convergence speed and quality for training image diffusion models by aligning their internal representations with pretrained, high quality external visual features, it is not designed for VDMs as it does not reflect consistency in temporal semantics during the alignment. To address this, we propose \textit{Cross-frame Representation Alignment} (CREPA), a novel regularization method for VDMs that explicitly incorporates temporal context by regularizing hidden states of a frame toward external pretrained features of adjacent frames, rather than only its own frame. Empirical evaluations show that CREPA significantly enhances visual fidelity and temporal consistency when fine-tuning large-scale video diffusion models such as CogVideoX-5B and Hunyuan Video in 3,000 training iterations using parameter-efficient approaches such as LoRA. We further validate our method by fine-tuning and evaluating on diverse datasets containing various visual attributes, demonstrating its effectiveness.
% \end{abstract}

%% file: sections/1_intro.tex
\section{Introduction}

Fine-tuning Video Diffusion Models (VDMs) to generate videos with specific attributes that are not expressible with text prompts presents a unique challenge in the field of generative AI. Especially, one may have a set of training videos with desired attributes, such as a specific cartoon style, physical attribute, or a scene staticity, to fine-tune and to generate videos accordingly. However, it is challenging to fine-tune large-scale VDMs at the user level, where limited training steps are often inevitable. This limitation often leads to suboptimal adaptation to training data, making it difficult for the model to capture nuanced attributes or styles present in the training data. Therefore, there is a growing need for effective yet efficient fine-tuning strategies that can bridge the gap between high computational requirements and user-level constraints, while preserving the rich generative capabilities of VDMs. However, this problem remains underexplored with few studies addressing this challenge.

Meanwhile, recent advancements such as Representation Alignment~\cite{yu2024representation} (REPA) significantly accelerate the convergence of image diffusion models. Specifically, REPA aligns the internal representations of Diffusion Transformers~\cite{peebles2023scalable} (DiT) with external visual features from self-supervised, pre-trained encoders such as DINOv2~\cite{oquabdinov2}. REPA builds on the insight that DiT behave like a Denoising Autoencoders~\cite{xiang2023denoising} (DAE), where the earlier transformer blocks behave as an encoder to process the noisy images, while the decoder predicts the noise based on the encoded features. Since DAE can learn less information as its input is inherently noisy, REPA enhances the intermediate hidden states of the DiT encoder by projecting and assimilating them to their corresponding pretrained features via an additional regularization objective. This simple yet effective distillation-based regularization method achieves over 17.5× faster convergence on SiT~\cite{ma2024sit} and improved image generation quality on ImageNet~\cite{deng2009imagenet}. Inspired by the observation, we first propose REPA*, a straightforward method of applying REPA to VDMs, and empirically found out that REPA could also benefit finetuning VDMs in some extent. 

However, we learned that the inherent nature of DAEs that extract hidden states from \textit{noisy} inputs lead to suboptimal convergence toward the \textit{sequence} of pretrained features. For instance, while it may be feasible to align noisy hidden states to each of their frame's pretrained feature individually, they can still be projected to arbitrary locations relative to its pretrained feature due to the noise in the input of the diffusion model. Such stochasticity can thus lead to semantic inconsistency of hidden states across frames during alignment. For instance, in Fig.~\ref{fig:teaser}, while REPA* shows improved convergence compared to model with no regularization (Vanilla) by better capturing the text prompt and the corresponding visual attributes in the training data compared to the model fine-tuned with score-prediction objective only, the generated frames often suffer from unnatural temporal context, such as broken transitions and objects fragmenting over time. This observation motivated our hypothesis that per-frame alignment does not ideally consider how the hidden states should semantically relate across frames.
% is not designed to improve relations of hidden states across the frames.
% fails to constrain how the hidden states relate across frames.

In this paper, we argue that further aligning hidden states of a frame using the pretrained features of adjacent frames can mitigate the problem and enhance the quality of generated videos. To achieve such alignment, we propose a novel distillation-based regularization method, which we refer to as \textit{Cross-frame Representation Alignment} (CREPA)~\footnote{We will release the code upon publication.}. Through empirical studies, we demonstrate that CREPA enables successful fine-tuning of large-scale video diffusion models such as CogVideoX-5B~\cite{yang2024cogvideox} and Hunyuan Video~\cite{kong2024hunyuanvideo} within only 3,000 training iterations, which amount to 9 and 13 hours of training time on a single A100 GPU respectively, using parameter-efficient fine-tuning techniques such as LoRA~\cite{hu2022lora}. We compare our method against vanilla fine-tuning with a score-matching objective alone, as well as fine-tuning with additional regularization methods such as REPA*, across multiple datasets exhibiting diverse characteristics, including unique visual styles, specific physical interactions, and 3D spatial consistency.

%% file: sections/2_related_works.tex
\section{Related Works}

\textbf{Video Diffusion Models} \ \ 
Recent advances in video generation have been driven by extending diffusion models to the video domain. Early works such as Align-Your-Latents~\cite{blattmann2023align}, Stable Video Diffusion~\cite{blattmann2023stable}, and Open-Sora~\cite{zheng2024open} laid the foundation for video diffusion models by adopting a U-Net architecture and incorporating disentangled spatial and temporal attention mechanisms, demonstrating large-scale training on video dataset. Subsequent large-scale models like CogVideoX~\cite{yang2024cogvideox}, HunyuanVideo~\cite{kong2024hunyuanvideo}, and Wan2.1~\cite{wang2025wan} leveraged DiT~\cite{peebles2023scalable} and scaled up to billions of parameters, enabling longer and more complex video sequences. These models introduced joint spatio-temporal attention modules to better capture temporal dependencies and enhance generative expressiveness. 

While these works primarily focus on architectural design and large-scale training, other lines of work demonstrate the utility of fine-tuning VDMs for task-specific applications. For instance, CustomTTT~\cite{bi2025customttt} and JointTuner~\cite{chen2025jointtuner} fine-tune VDMs to enable object-centric video generation with controllable motion and appearance. In the meantime, Long Context Tuning~\cite{guo2025long} adapts VDMs for long-range temporal generation by additionally modifying attention mechanisms to enhance multi-shot consistency. However, since none of these methods explicitly target the enhancement of semantic consistency through regularization as an auxiliary optimization objective during fine-tuning, we believe our approach can complement prior and future works that fine-tunes VDMs.

\textbf{Fine-Tuning Large-Scale Models} \ \ The specialization of large-scale models through fine-tuning or adaptation has become a widely adopted strategy for tailoring pre-trained models to specific tasks or domains. This paradigm leverages the broad generalization capabilities of foundation models while enabling efficient task-specific refinement. Numerous methods have been introduced to facilitate this process across different modes of data, including large language models~\cite{brown2020language, gururangan2020don}, vision-language models~\cite{liu2023visual, liu2024improved}, image diffusion models~\cite{ruiz2023dreambooth, shah2024ziplora}, vision foundation models~\cite{hu2023efficiently, yue2024improving}, and even in geometric foundation models~\cite{lu2024lora3d}. These methods often leverage parameter-efficient fine-tuning methods such as LoRA~\cite{hu2022lora} to maximally leverage prior knowledge of pretrained model and maximize training efficeincy.

It is important to recognize that fine-tuning strategies often reflect the intrinsic characteristics of the data modality. For example, fine-tuning text-to-image diffusion models typically focuses on disentangling visual style from object-centric content~\cite{shah2024ziplora, wang2023stylediffusion}. In the case of 3D foundation models such as DUSt3R~\cite{wang2024dust3r}, adaptation methods are designed to accurately estimate per-pixel confidence in depth back-projection to ensure reliable geometric reasoning~\cite{lu2024lora3d}. Analogously, we posit that fine-tuning VDMs should explicitly account for semantic consistency, a core property of video data that is not relevant to models designed for other data domains.

%% file: sections/3_prelim.tex
\section{Preliminaries}
Diffusion models~\cite{sohl2015deep, ho2020denoising, song2020denoising, song2020score} are a class of generative models that synthesize data by inverting a gradual noising process applied to clean samples. This procedure consists of two main phases: the forward process and the reverse process.
\vspace{-0.3cm}
\paragraph{Forward Process.}
The forward process gradually transforms a data sample $x_0 \sim p(x_0)$ into a noisy variable $x_T$ through a sequence of perturbations, typically by adding Gaussian noise. This transformation can be modeled either in continuous time using a stochastic or ordinary differential equation:
\begin{equation}
    dx = f(x, t) dt + g(t) dw,
\end{equation}
where $f(x, t)$ defines the deterministic drift, $g(t)$ controls the noise magnitude, and $dw$ denotes a standard Wiener process (for SDE~\cite{song2020score}) or is omitted in the deterministic case (for ODE~\cite{song2020denoising}). As $t \to T$, the sample $x_t$ approaches a known prior distribution, such as $\mathcal{N}(0, I)$.
\vspace{-0.3cm}
\paragraph{Reverse Process.}
The generative process aims to reconstruct samples from the data distribution by reversing the forward dynamics. Assuming access to the data score function $\nabla_x \log p_t(x)$, the reverse-time dynamics can be expressed as:
\begin{equation}
    dx = \left[ f(x, t) - g(t)^2 \nabla_x \log p_t(x) \right] dt + g(t) d\bar{w},
\end{equation}
where $d\bar{w}$ denotes the reverse-time Wiener process. In practice, the score function is approximated by a neural network $s_\theta(x, t)$ trained to match $\nabla_x \log p_t(x)$ via denoising score matching. Sampling is then performed by numerically solving the reverse-time SDE or its deterministic counterpart.
\vspace{-0.3cm}
\paragraph{Denoising Score Matching Loss.}
To train the score-based model $s_\theta(x, t)$, one commonly minimizes a denoising score matching loss, which encourages the predicted score to match the true gradient of the log-density of $x_t$. A widely used objective in diffusion models is the simplified variational bound:
\begin{equation}
    \mathcal{L}_{\text{score}}(\theta) = \mathbb{E}_{x_0, \epsilon, t} \left[ \left\| \epsilon - \epsilon_\theta(x_t, t) \right\|_2^2 \right],
\end{equation}
where $\epsilon_\theta$ denotes a neural network trained to predict the noise added at time $t$, and $x_t$ is generated from $x_0$ using the forward noising process. This formulation is equivalent to score matching under certain assumptions and is widely adopted for its stability and empirical performance.

%% file: sections/4_methods.tex
\begin{figure}[t!]
    \centering
    \includegraphics[width=\textwidth]{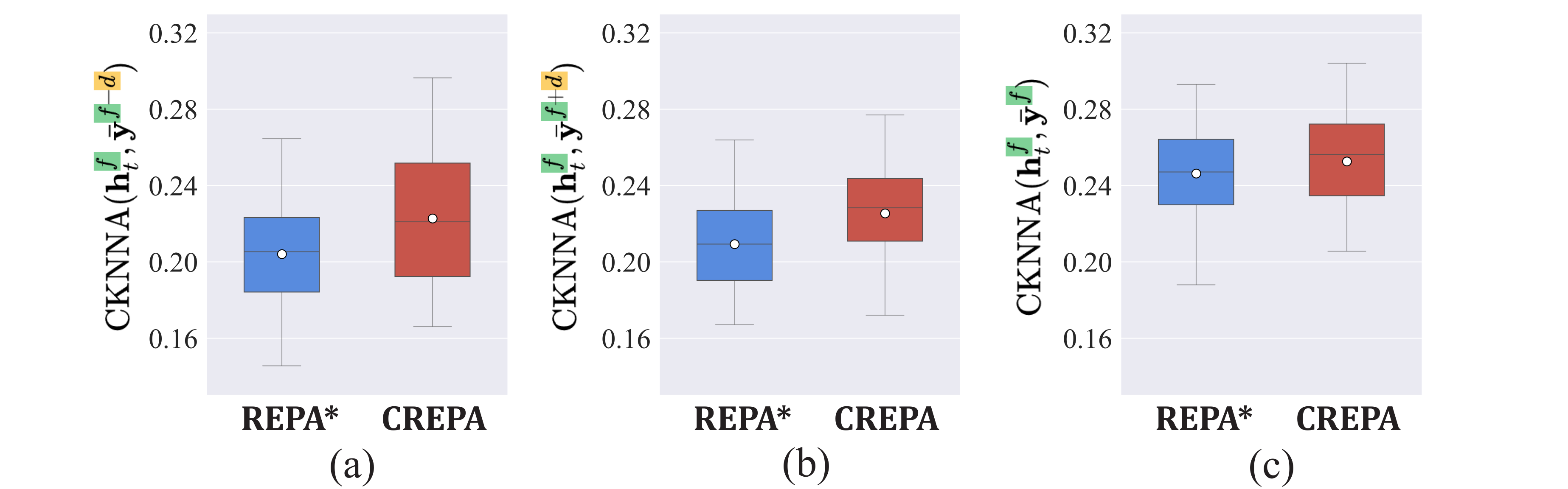}
    \caption{\textbf{CKNNA~\cite{huh2024platonic} between hidden states of a frame to pretrained features}, or representation alignment, to (a) preceding, (b) future, and (c) current frames. CREPA promotes alignment to adjacent frames, while maintaining or even slightly improving the alignment to the current frame.}
    \label{fig:pullback}
\end{figure}

\section{Methods}
\vspace{-0.2cm}
We begin by proposing REPA*, an extension of REPA for fine-tuning VDMs, and make empirical verification on its limitations of disregarding the semantics of past and future frames during the alignment of noisy hidden states. To address this issue, we introduce CREPA, which explicitly incorporates temporal context through pretrained features of adjacent frames.

\subsection{Per-frame Representation Alignment for Video Diffusion Models}
\label{subsec:repa}
\vspace{-0.2cm}
Recall that REPA aligns the internal representations of Image Diffusion Transformers with external visual features obtained from self-supervised, pre-trained encoders. Motivated by the observation that DiTs function similarly to Denoising Autoencoders~\cite{xiang2023denoising}, where early transformer blocks act as an encoder to extract relevant information from noisy image and later blocks serve as a decoder, REPA enhances noisy hidden states of DiT encoders by projecting and supervising them with corresponding pretrained features that contain richer information for distillation. Correspondingly, better and faster convergence introduced by REPA can also benefit the fine-tuning of DiT on user-level where number of training iterations is often limited.

We first propose REPA*, an extension to per-frame application of REPA in VDMs. Since a general, widely used pretrained visual encoder for videos does not exist or focus only on specific visual domain such as Minecraft~\cite{baker2022vpt}, we instead assume a pre-trained image encoder for the method. Specifically, given a frame $x^{f}_{0}$ of a clean video $x_0 \sim p(x_0)$ and a pre-trained encoder $E$, we first extract per-frame pretrained feature $\bar{\textbf{y}}^{f}=E(x^{f}_{0})$. Then, the DiT encoder $g_{\theta}$, which is composed of the first few DiT blocks, encodes the noisy video input $x_t$ to the hidden state $\textbf{h}_t = g_{\theta}(x_t, t)$, where per-frame hidden state $\textbf{h}^{f}_t$ is then projected to the pretrained feature space $\mathcal{\bar{Y}}$ via small MLP $h_{\phi}(\cdot)$ to be assimilated with its corresponding pretrained feature via the regularization-based distillation:

\begin{equation}
\label{eq:repa}
\mathcal{L}_{\text{align}}(\phi) := - \mathbb{E}_{\textbf{x}_0, \epsilon, t} \left[ \sum_{f} \text{sim}(\bar{\textbf{y}}^{f}, h_{\phi}(\textbf{h}^{f}_t)) \right],
\end{equation}

where $\text{sim}(\cdot, \cdot)$ is a similarity function. Thus, the final objective for fine-tuning becomes 

\begin{equation}
    \mathcal{L} := \mathcal{L}_{\text{score}} + \lambda \mathcal{L}_{\text{align}},
\end{equation}

where $\lambda$ is a hyperparameter to control the strength of the alignment.

\begin{figure}[t!]
    \centering
    \includegraphics[width=\textwidth]{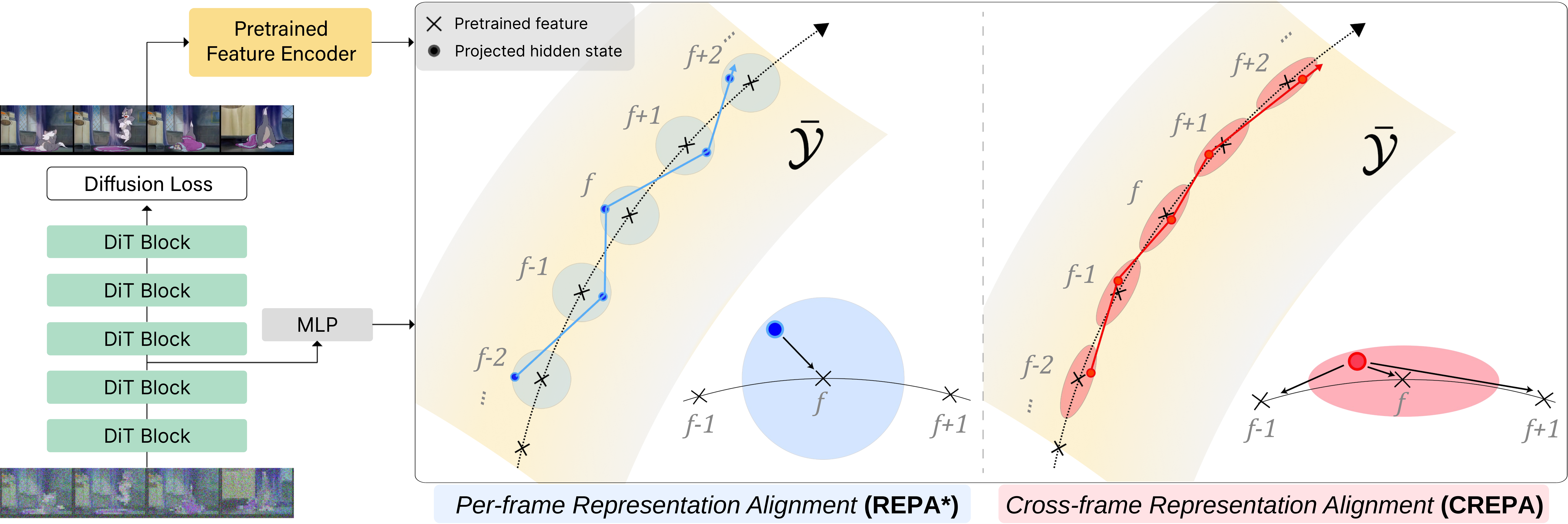}
    \caption{\textbf{Overview of CREPA and comparison to REPA*}. By aligning hidden states not only to the current but also to the adjacent pretrained features, CREPA further guides the hidden state representations toward the temporal manifold formed by the sequence of pretrained features.}
    \label{fig:overview}
\end{figure}

\subsection{Cross-frame Representation Alignment}
\label{subsec:crepa}
\vspace{-0.2cm}
However, due to the inherent nature of DAEs training on noisy inputs and jointly training $h_{\theta}$ during fine-tuning, we learned that regularizing VDMs with Eq.~\eqref{eq:repa} alone cannot prevent the projected hidden state for each frame from converging to arbitrary locations with respect to their corresponding pretrained feature. We hypothesize that such ill-posed solutions cause weaker \textit{cross-frame representation alignment}, which we define as a semantic similarity between the hidden states of a frame and pretrained features of its adjacent frames. We argue that weaker cross-frame representation alignment causes semantically inconsistent and suboptimal generation in VDMs.

\textbf{Empirical observation} \ \ We first quantify the degree of cross-frame representation alignment for models fine-tuned with REPA*, and compare with CREPA. Specifically
for each model, we compute the similarity between the hidden state of a current frame and the pretrained features of both the adjacent and current frames. For similarity, we utilize CKNNA~\cite{huh2024platonic}, a kernel alignment metric that extends CKA~\cite{kornblith2019similarity} by incorporating mutual nearest neighbors to compare representations across different feature spaces. We follow REPA~\cite{yu2024representation} for its implementation, provide further details in Appendix~\ref{app:analysis}, and report the results in Fig.~\ref{fig:pullback}. Compared to REPA*, CREPA yields higher similarity between the current hidden states and adjacent pretrained features, as shown in Fig.~\ref{fig:pullback}-(a),(b), while maintaining or slightly improving the alignment with the pretrained features of the current frame in Fig.~\ref{fig:pullback}-(c). This suggests that regularizing hidden states solely based on the current frame is under-constrained: \textit{noisy hidden states, which can converge to different locations with respect to its corresponding pretrained features, may exhibit arbitrary distances to the pretrained features of adjacent frames while satisfying similar distance to that of the current frame.}

\textbf{Our approach} \ \ To address such under-constraint problem caused by the nature of DAEs encoding noisy input, we draw inspiration from total variation regularization, a well-known technique that penalizes noisy variations using information from neighboring pixels in raster space~\cite{rudin1992nonlinear, yeh2022total, mahendran2015understanding}. Analogously, we encourage to reduce the uncertainty of alignment in temporal dimension by using information of pretrained features from adjacent frames. Note that our objective extends beyond mere temporal smoothness; instead of directly smoothing among the inherently noisy sequence of hidden states, we impose cross-frame distillation from adjacent pretrained features, where semantic is more robustly captured. In other words, our objective is to align hidden states with the semantic evolution of pretrained features, ensuring that hidden states of a frame are further regularized toward the temporal semantic context present in training data. To do so, we define a simple yet novel regularization objective for alignment:
\begin{equation}
\label{eq:crepa}
\mathcal{L}_{\text{align}}(\phi) := - \mathbb{E}_{\textbf{x}_0, \epsilon, t} \left[ \sum_{f} \left( \text{sim}(\bar{\textbf{y}}^{f}, h_{\phi}(\textbf{h}^{f}_t)) + \sum_{k \in \mathcal{K}} e^{-\frac{|k-f|}{\tau}} \cdot  \text{sim}(\bar{\textbf{y}}^{k}, h_{\phi}(\textbf{h}^{f}_t)) \right) \right],
% ^{*}\textbf{h}^{f}_{t}, ^{*}h_{\phi}(\cdot)  = \argmax_{\textbf{h}^f_t, h_{\phi}} \mathbb{E}_{\textbf{x}_0, \epsilon, t} \left[ \sum_{f} \left( \text{sim}(\bar{\textbf{y}}^{f}, h_{\phi}(\textbf{h}^{f}_t)) + \sum_{k \in \mathcal{K}} e^{-\frac{|k-f|}{\tau}} \cdot  \text{sim}(\bar{\textbf{y}}^{k}, h_{\phi}(\textbf{h}^{f}_t)) \right) \right],
\end{equation}
where $\mathcal{K} = \{ f-d, f+d \}$, $d$ is the adjacency parameter, and $\tau$ is the temperature coefficient. A conceptual overview of our method and comparison to REPA* is illustrated in Fig.~\ref{fig:overview}.

%% file: sections/5_experiments.tex
\begin{table*}[!t]
\centering
\resizebox{\textwidth}{!}{
\begin{tabular}{l|ccccc}
\toprule
\textbf{Models}  & Aesthetic Quality $\uparrow$  & Background Consistency $\uparrow$ & Motion Smoothness $\uparrow$  & Subject Consistency $\uparrow$ & Imaging Quality $\uparrow$ \\ 
\midrule
Vanilla                       & 0.5013             & 0.9293             & 0.9827    & 0.8784     & 0.5982                                     \\ 
REPA*                         & 0.5145             & 0.9303             & 0.9828    & 0.8798     & \textbf{0.6284}\cellcolor[HTML]{DAE8FC}                            \\ 
CREPA (\textit{Ours})         & \textbf{0.5207}\cellcolor[HTML]{DAE8FC}    & \textbf{0.9351}\cellcolor[HTML]{DAE8FC}    & \textbf{0.9847}\cellcolor[HTML]{DAE8FC} & \textbf{0.8891}\cellcolor[HTML]{DAE8FC} & 0.6147                        \\ 
\midrule
Vanilla & 0.5243   & 0.9347  & 0.9809   &   0.9107   & 0.6285   \\ 
REPA*      & \textbf{0.5362} \cellcolor[HTML]{DAE8FC} & 0.9362  & 0.9849     &   0.8984  & 0.6280  \\ 
CREPA (\textit{Ours})  & 0.5351     & \textbf{0.9491} \cellcolor[HTML]{DAE8FC}   &  \textbf{0.9895} \cellcolor[HTML]{DAE8FC}   & \textbf{0.9205} \cellcolor[HTML]{DAE8FC} & \textbf{0.6451} \cellcolor[HTML]{DAE8FC}       \\ 

\bottomrule
\end{tabular}
}

\caption{\textbf{VBench~\cite{huang2024vbench} evaluation on Hunyuan Video~\cite{kong2024hunyuanvideo}} \textit{(Top)} \textbf{and CogVideoX-5B} \textit{(Bottom)}. CREPA outperforms baselines on most metrics, especially on metrics related to semantic consistency.}
\vspace{-4mm}
\label{tab:vbench_hunyuan}
\end{table*}

\section{Experiments}
\vspace{-0.2cm}
\subsection{Setup}
\textbf{Models} \ \ We employ CogVideoX-5B~\cite{yang2024cogvideox}, which features an expert transformer with adaptive LayerNorm and 3D full attention for effective fusion of text and video features. We also employ Hunyuan Video~\cite{kong2024hunyuanvideo}, a dual-to-single stream transformer designed with a unified attention mechanism with 13B parameters. We also experiment with the Image-to-Video model of CogVideoX-5B.

\textbf{Baselines} \ \ Since few works have addressed convergence and semantic consistency in VDM fine-tuning, we establish and compare against our own baselines. Specifically, we compare CREPA with Vanilla, models fine-tuned with score prediction objective only, and REPA*.

\textbf{Implementation Details} \ \ We use $\lambda = 0.5$ for CogVideoX-5B and $\lambda = 1$ for Hunyuan video, where we empirically found to work the best on REPA*. We also use $d=1$ and $\tau=1$ for CREPA. As both models define DiT models in the latent space formed by VAE that introduces $4\times$ temporal compression rate, the effective adjacency for $d=1$ is $\leq 4$. For pretrained models, we used DINOv2-g~\cite{oquabdinov2} following REPA~\cite{yu2024representation}. To determine the hidden state layer for alignment in DiT, we first perform linear probing to identify the encoder, followed by a layer-wise analysis to select the optimal layer, as detailed in Appendix~\ref{app:linear_probing}. Based on this procedure, we select the 8th layer for CogVideoX and the 10th layer for Hunyuan Video. We train with a single A100 NVIDIA GPU with 80G VRAM for $3000$ iterations for all models.  

\textbf{Dataset} \ \ We experimented on 7 video datasets that contains various visual and temporal attributes. We list the datasets into 4 categories based on their characteristics:
\begin{itemize}[leftmargin=*, noitemsep, topsep=0cm]
  \item \textbf{Tom and Jerry}~\cite{wildheart2024tomjerry} and \textbf{Disney}~\cite{wildheart2024disney} consist of clips of cartoons with unique illustrative styles.
  \item \textbf{Crush}~\cite{finetrainers_crush_smol}, \textbf{Cakeify}~\cite{finetrainers_cakeify_smol} and  \textbf{Squish}~\cite{finetrainers_squish_pika} consist of curated video clips with specific physical interactions for VDMs to adapt to. Specifically, Crush describes an object flattened under a hydraulic press, Cakeify describes a cake designed to not look like a cake being cut like a cake, and Squish describes an object being squished as if it is made out of a soft material. 
  \item \textbf{DL3DV}~\cite{ling2024dl3dv} contains video captures of bounded and unbounded scenes with no dynamic objects included. As so, DL3DV is widely used for Novel View Synthesis and 3D reconstruction tasks. Out of $\approx 10K$ video clips, we randomly sample 200 videos for train and test each. We use CogVLM2-Video~\cite{hong2024cogvlm2} to generate captions.
  \item \textbf{Scenes}~\cite{bigdata-pw_scenes} contains clips of a movie, which is photorealistic yet contains a specific visual nuance. 
\end{itemize}

\textbf{Evaluation} \ \ We evaluate our method on two representative aspects of video generation: perceptual quality and semantic consistency. For perceptual-level evaluation, we report Fréchet Video Distance (FVD) and Inception Score (IS) using the CogVideoX-5B I2V model finetuned with the DL3DV dataset~\cite{ling2024dl3dv}. FVD is computed using an I3D model , and IS is measured with a C3D model. These metrics reflect the distributional alignment and frame-level realism of generated videos. 

To assess frame-wise semantic consistency and visual coherence, we adopt VBench~\cite{huang2024vbench}, a multi-attribute benchmark. Criteria relevant to semantic consistency in VBench are Motion Smoothness, Background Consistency, Subject Consistency. We also measure Aesthetic Quality and Imaging Quality in VBench to further evaluate the perceptual quality.

\subsection{Results}
\label{subsec:results}
\vspace{-0.2cm}

\textbf{Quantitative Results} 
We compare CREPA with Vanilla and REPA on standard benchmarks using VBench, as shown in Table~\ref{tab:vbench_hunyuan}. CREPA consistently outperforms both baselines across most metrics in VBench. It shows strong semantic consistency (e.g., background and subject alignment) and smooth motion, maintaining coherence throughout generated videos. While REPA is on par with a few categories related to perceptual quality, CREPA offers more balanced and robust performance including metrics on semantic consistency, highlighting its effectiveness in generating high-quality and semantically consistent video.

We report FVD and IS on Table~\ref{table:fvd}. CREPA shows reasonable improvements over the baselines, achieving the best performance in both metrics. This indicates that CREPA excels in both distribution-level similarity to real videos and frame-level object quality.

% Furthermore, in the novel view synthesis task designed to assess spatial consistency (Table~\ref{tab:3dgs}), CREPA demonstrates clear advantages. Under the 3DGS reconstruction-based evaluation of both T2V and I2V generation, CREPA outperforms the baselines across all key metrics, including PSNR, SSIM, and LPIPS. These results strongly suggest that CREPA generates videos with superior spatial and geometric consistency.

% Overall, the results indicate that CREPA best captures the underlying data distribution and achieves more effective training compared to the baseline models, leading to higher-quality generation.

\begin{figure}[t!]
    \centering
    \includegraphics[width=\textwidth]{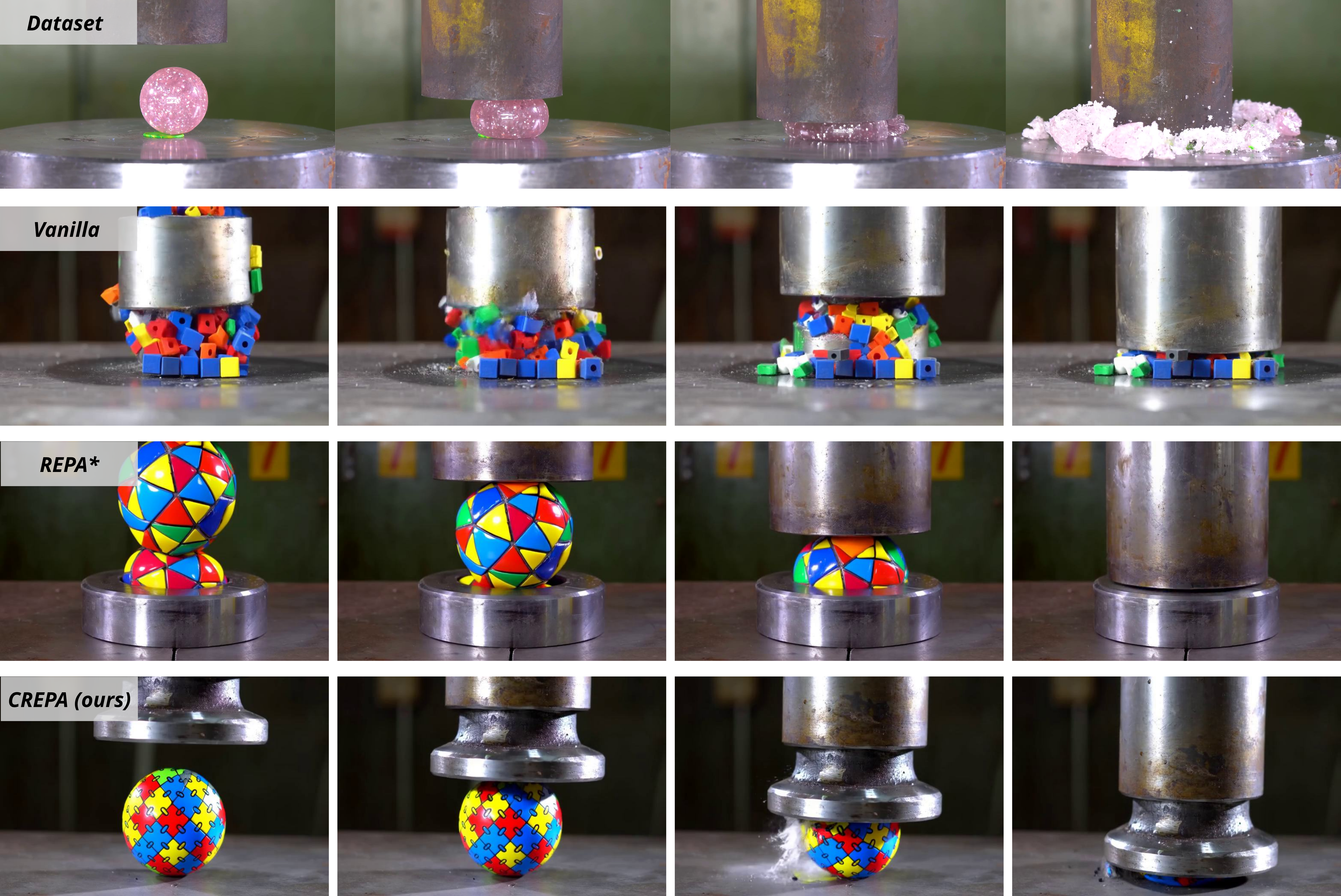}
    \caption{\textbf{Videos generated by Hunyuan Video~\cite{kong2024hunyuanvideo} fine-tuned on Crush~\cite{finetrainers_crush_smol} dataset.} CREPA enhances convergence relative to Vanilla by better learning the physical attribute of the data. Also, CREPA yields better semantic consistency compared to REPA*. Text prompt reported in Appendix~\ref{app:textgen}.}
    \label{fig:main_crush}
\end{figure}

\begin{figure}[t!]
    \centering
    \includegraphics[width=\textwidth]{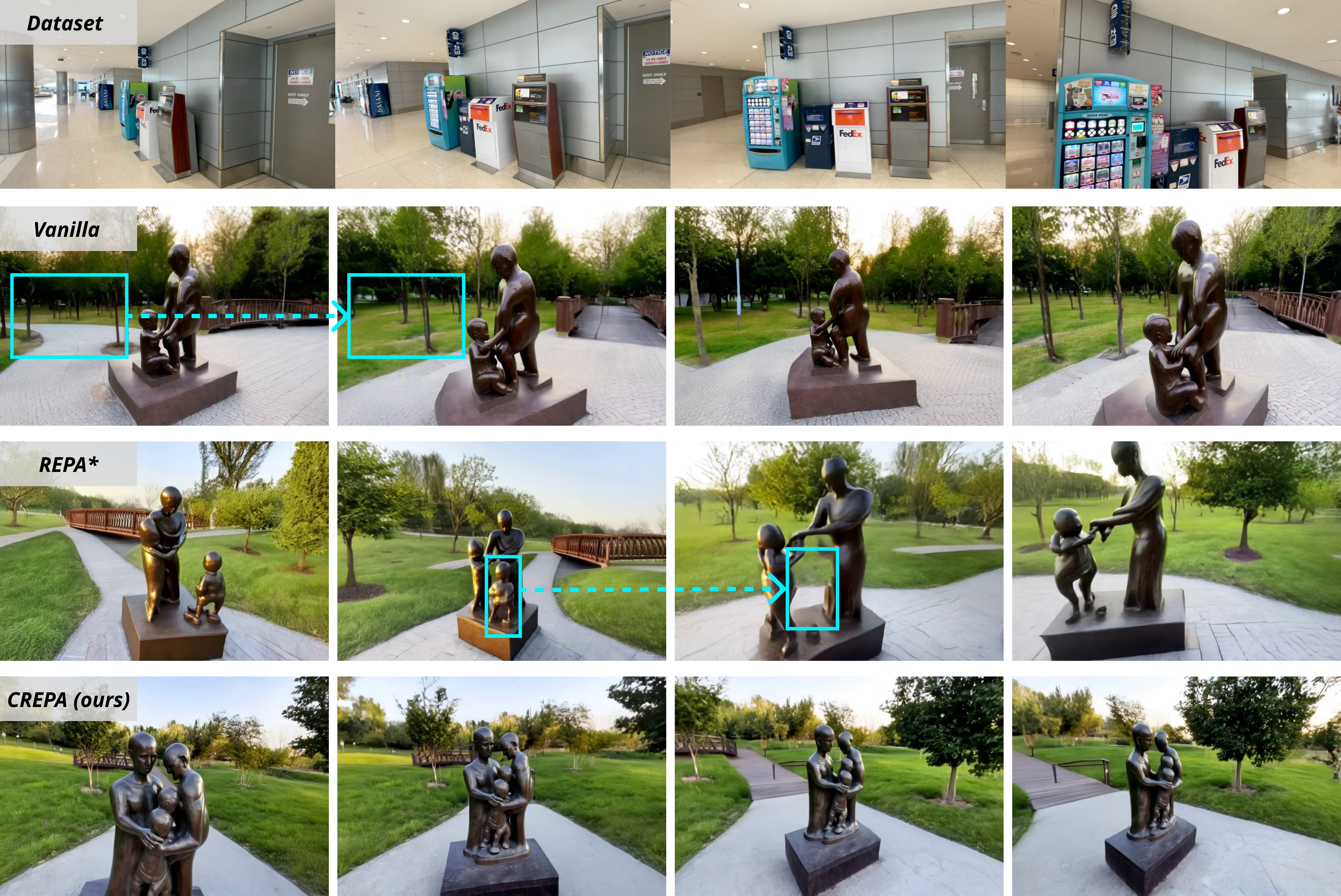}
    \caption{\textbf{Videos generated by CogVideoX-5B~\cite{yang2024cogvideox} fine-tuned on DL3DV~\cite{ling2024dl3dv} dataset.} Compared to REPA*, CREPA yields more semantically consistent objects in video across the frames. Meanwhile, Vanilla yields disappearing road in the background. Text prompt reported in Appendix~\ref{app:textgen}.}
    \vspace{-0.15cm}
    \label{fig:main_dl3dv}
\end{figure}

\begin{figure}[t!]
    \centering
    \includegraphics[width=\textwidth]{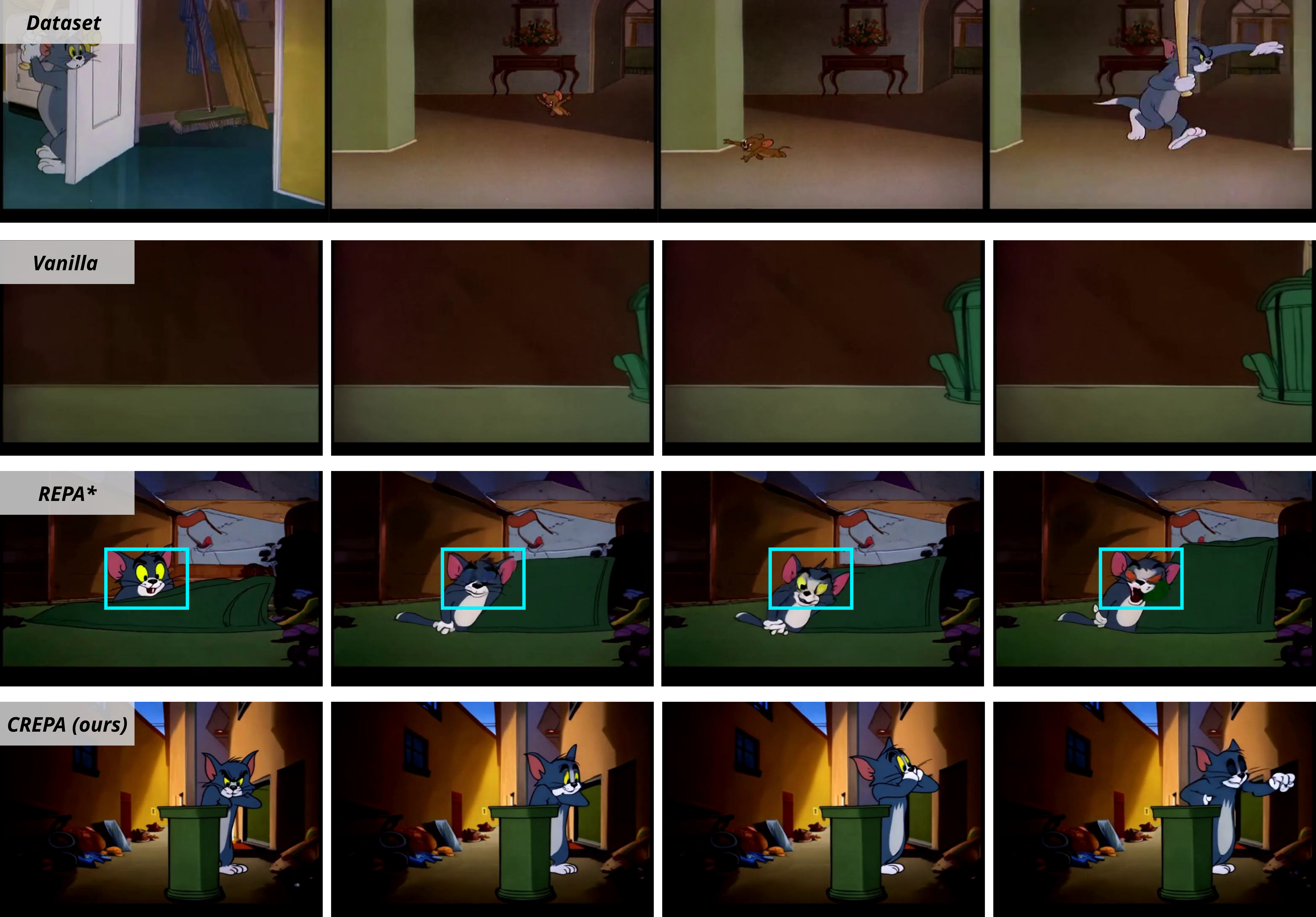}
    \caption{\textbf{Videos generated by CogVideoX-5B~\cite{yang2024cogvideox} fine-tuned on Tom and Jerry~\cite{wildheart2024tomjerry} dataset.} CREPA improves convergence and semantic consistency over Vanilla and REPA*, respectively. Text prompt reported in Appendix~\ref{app:textgen}.}
    \label{fig:main_tom}
\end{figure}

\textbf{Qualitative Results}\footnote{We strongly recommend to refer to the video results in the project page.} \ \ We report qualitative results on the baselines as well as an example from training data as a reference. Notably in Fig.~\ref{fig:main_crush} and Fig.~\ref{fig:main_tom}, both REPA* and CREPA shows better generation quality compared to Vanilla, demonstrating that the distillation-based regularizations do facilitate better convergence. However, REPA* often yields semantically inconsistent videos, (i.e., shape and appearance of statue changes over frames in Fig.~\ref{fig:main_dl3dv}, the shape of the ball changes in physically implausible ways in Fig.~\ref{fig:main_crush}, or the appearance of the character changes in Fig.~\ref{fig:main_tom}). These observations suggest that CREPA is more effective at preserving semantic consistency present in training data. We report additional results in Appendix~\ref{app:more}.

% 더 추가적인 qualitative 피규어는 appendix X에 있습니다. 

% \begin{figure}[t!]
%     \centering
%     \includegraphics[width=\textwidth]{figures/video_samples/Hun_cakeify.pdf}
%     \caption{Finetuning dataset: cakeify-smol}
%     \label{fig:hun_cakeify}
% \end{figure}

% \begin{figure}[t!]
%     \centering
%     \includegraphics[width=\textwidth]{figures/video_samples/Hun_crush.pdf}
%     \caption{Finetuning dataset: crush}
%     \label{fig:hun_cakeify}
% \end{figure}

% \begin{figure}[t!]
%     \centering
%     \includegraphics[width=\textwidth]{figures/video_samples/Hun_disney.pdf}
%     \caption{Finetuning dataset: disney}
%     \label{fig:hun_cakeify}
% \end{figure}

% \begin{figure}[t!]
%     \centering
%     \includegraphics[width=\textwidth]{figures/video_samples/Hun_dl3dv.pdf}
%     \caption{Finetuning dataset: dl3dv}
%     \label{fig:hun_cakeify}
% \end{figure}

% \begin{figure}[t!]
%     \centering
%     \includegraphics[width=\textwidth]{figures/video_samples/Hun_scenes.pdf}
%     \caption{Finetuning dataset: scenes}
%     \label{fig:hun_cakeify}
% \end{figure}

% \begin{figure}[t!]
%     \centering
%     \includegraphics[width=\textwidth]{figures/video_samples/Hun_squish.pdf}
%     \caption{Finetuning dataset: squish}
%     \label{fig:hun_cakeify}
% \end{figure}

% \begin{figure}[t!]
%     \centering
%     \includegraphics[width=\textwidth]{figures/video_samples/Hun_tom.pdf}
%     \caption{Finetuning dataset: tom and Jerry}
%     \label{fig:hun_cakeify}
% \end{figure}

\begin{figure}[t!]
    \centering
    \includegraphics[width=\textwidth]{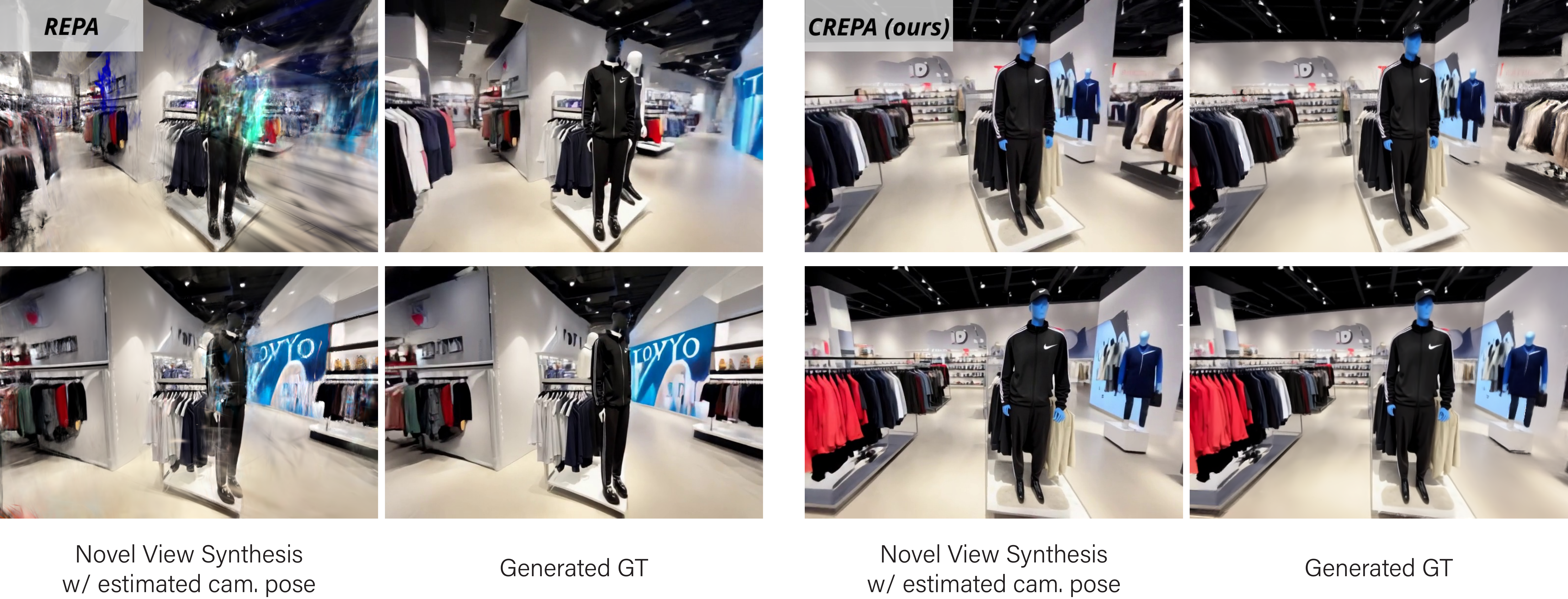}
    \caption{\textbf{Novel view synthesis on 3D scenes reconstructed with generated videos.} Spatially inconsistent videos cause inaccurate camera pose estimation from COLMAP~\cite{schoenberger2016sfm}, as well as imprecise supervision for rendering loss while training 3DGS~\cite{kerbl20233d}.}
    \label{fig:3dgs}
\end{figure}

\textbf{Application to Novel View Synthesis} \ \ VDMs have recently been adapted into World Fondational Models (WFMs), which are essentially video generative models for simulating 3D consistent scene navigation~\cite{agarwal2025cosmos, liang2024wonderland, hecameractrl}. They are often fine-tuned on spatially consistent scene navigation dataset such as DL3DV~\cite{ling2024dl3dv}. We hypothesize that CREPA is also suitable for enhancing WFMs by better learning 3D spatial consistency inherent in inter-frame relations within DL3DV. To test this, we first use COLMAP~\cite{schoenberger2016sfm} to estimate camera poses of video frames generated from CogVideoX-5B models, each of which are finetuned on DL3DV under the baselines and our method. Then, we reconstruct 3D scenes with 3DGS~\cite{kerbl20233d} using all but every 8th frame, and perform novel view synthesis (NVS) on these held-out views. Since NVS relies solely on other frames for rendering, higher similarity between the image rendered by NVS and the original frame at the target view generated by the VDM indicates stronger 3D consistency in the generated video. As Fig.~\ref{fig:3dgs} and Table.~\ref{tab:3dgs} shows, CREPA yields better NVS performance compared to REPA* and vanilla.

% \paragraph{Novel view synthesis via neural 3D reconstruction}

% \begin{table*}[h]
% \centering
% \resizebox{0.7\textwidth}{!}{
% \begin{tabular}{l|ccc|ccc}
% \toprule
% & \multicolumn{3}{c|}{\textbf{T2V}} & \multicolumn{3}{c}{\textbf{I2V}} \\
% \textbf{Models}  & PSNR $\uparrow$ & SSIM $\uparrow$ & LPIPS $\downarrow$  & PSNR $\uparrow$ & SSIM $\uparrow$  & LPIPS $\downarrow$ \\ 
% \midrule
% Vanilla & 21.83    & 0.741  & 0.255    & 22.10    & 0.755    & 0.250                  \\ 
% REPA   & 22.17 & 0.742 &  0.253    & 22.45    & 0.758    & 0.252                  \\ 
% CREPA (\textit{Ours})  & \textbf{22.66} \cellcolor[HTML]{DAE8FC} & \textbf{0.760} \cellcolor[HTML]{DAE8FC} &  \textbf{0.248} \cellcolor[HTML]{DAE8FC}  & \textbf{22.88} \cellcolor[HTML]{DAE8FC} & \textbf{0.770} \cellcolor[HTML]{DAE8FC} & \textbf{0.245} \cellcolor[HTML]{DAE8FC} \\ 
% \bottomrule
% \end{tabular}
% }
% \caption{\textbf{Quantitative results on novel view synthesis.} Given models finetuned with DL3DV~\cite{ling2024dl3dv} dataset, we generate videos with T2V and I2V models, followed by estimating camera pose for each frame per video using COLMAP~\cite{schoenberger2016sfm}. Then, we train and evaluate 3DGS~\cite{kerbl20233d} using the camera pose estimations to measure spatial consistency from the generated videos.}
% % Crop해서 사용하는 REFace를 제외하고 모든 부분에서 모든baseline들을 이기며, 심지어 REFace와 comparable한 결과를 얻는다.
% \label{tab:3dgs}
% \end{table*}

% \documentclass{article}
% \usepackage{graphicx}
% \usepackage{booktabs}
% \usepackage[table]{xcolor}
% \usepackage{caption}

% \definecolor{pastelblue}{HTML}{DAE8FC}

% \begin{document}

\noindent
\begin{minipage}[t]{0.308\textwidth}
\vspace{0pt}
\centering
\resizebox{\textwidth}{!}{
\begin{tabular}{l|cc}
\toprule
% \multicolumn{3}{c}{\textbf{I2V}} \\
& \multicolumn{2}{c}{} \\
\textbf{Models} & FVD $\downarrow$ & IS $\uparrow$ \\
\midrule
Vanilla & 305.542 & 34.1 \\
REPA* & 291.388 & 35.2 \\
CREPA (\textit{Ours}) & \textbf{281.192} \cellcolor{pastelblue} & \textbf{35.8} \cellcolor{pastelblue} \\
\bottomrule
\end{tabular}
}
\vspace{0.5em}
\captionof{table}{\textbf{Quantitative results on FVD and IS.} \ \ CREPA yields better FVD and IS compared to the baselines, indicating its potential for better visual fidelity.}
\label{table:fvd}
\end{minipage}
\hfill
\begin{minipage}[t]{0.677\textwidth}
\vspace{0pt}
\centering
\resizebox{\textwidth}{!}{
\begin{tabular}{l|ccc|ccc}
\toprule
& \multicolumn{3}{c|}{\textbf{T2V}} & \multicolumn{3}{c}{\textbf{I2V}} \\
\textbf{Models} & PSNR $\uparrow$ & SSIM $\uparrow$ & LPIPS $\downarrow$ & PSNR $\uparrow$ & SSIM $\uparrow$ & LPIPS $\downarrow$ \\
\midrule
Vanilla & 21.83 & 0.741 & 0.255 & 22.10 & 0.755 & 0.250 \\
REPA* & 22.17 & 0.742 & 0.253 & 22.45 & 0.758 & 0.252 \\
CREPA (\textit{Ours}) & \textbf{22.66} \cellcolor{pastelblue} & \textbf{0.760} \cellcolor{pastelblue} & \textbf{0.248} \cellcolor{pastelblue} & \textbf{22.88} \cellcolor{pastelblue} & \textbf{0.770} \cellcolor{pastelblue} & \textbf{0.245} \cellcolor{pastelblue} \\
\bottomrule
\end{tabular}
}
\vspace{0.5em}
\captionof{table}{\textbf{Quantitative results on novel view synthesis.} Given models finetuned with DL3DV~\cite{ling2024dl3dv} dataset, we generate videos with T2V and I2V models, followed by estimating camera pose for each frame per video using COLMAP~\cite{schoenberger2016sfm}. Then, we train and evaluate 3DGS~\cite{kerbl20233d} using the camera pose estimations to measure spatial consistency from the generated videos.}
\label{tab:3dgs}
\end{minipage}

% \end{document}

\textbf{User Study} \ \ We additionally conducted a user study to evaluate the perceptual quality of the generated videos across six criteria: text-video alignment, video quality, motion quality, semantic consistency, training data reflectivity, and overall preference. Participants consistently favored videos generated with CREPA over those from REPA* and Vanilla across all aspects. Full details of the user study and results are provided in Appendix~\ref{app:user_study}.

%% file: sections/6_conclusion_and_future_works.tex
\section{Conclusion}

In this work, we proposed CREPA, a regularization method for fine-tuning VDMs. We first introduced REPA*, an adaptation of REPA to VDMs, for better and faster convergence for VDM fine-tuning. However, we empirically showed that aligning hidden states only to their corresponding frame via REPA* is insufficient to ensure cross-frame semantic consistency. CREPA addresses this limitation by additionally aligning hidden states to pretrained features from adjacent frames. Through extensive experiments across diverse datasets and models, we demonstrated that CREPA improves convergence compared to model fine-tuned without any regularization, while enhancing semantic consistency compared to the model trained with REPA*.

\textbf{Limitation} \ \ A key limitation of our work is that the proposed regularization method requires searching across DiT layers for different VDMs. However, this layer search does not need to be performed by every end user. A single representative or the model distributor can carry out the search once, and share the optimal layer index with downstream users, making the method practical for broader adoption.

% \textbf{Ethical Considerations} The enhanced generative capabilities enabled by our method may also increase the risk of misuse, such as in the creation of misleading or harmful content. This underscores the need for strict ethical guidelines and responsible dissemination practices in deploying such technologies.

\textbf{Future Works} \ \ Our distillation-based regularization technique presents an interesting opportunity for application during the pre-training phase of VDMs. However, we defer such exploration to future work, particularly by institutions with access to large-scale and high-quality video datasets and sufficient computational resources.

%% file: sections/7_appendix.tex
\newpage
\appendix
\section*{Appendix}
\section{Locating layers for hidden-state retrieval}
\label{app:linear_probing}
\begin{figure*}[!h]
    \centering
    \includegraphics[width=0.9\textwidth]{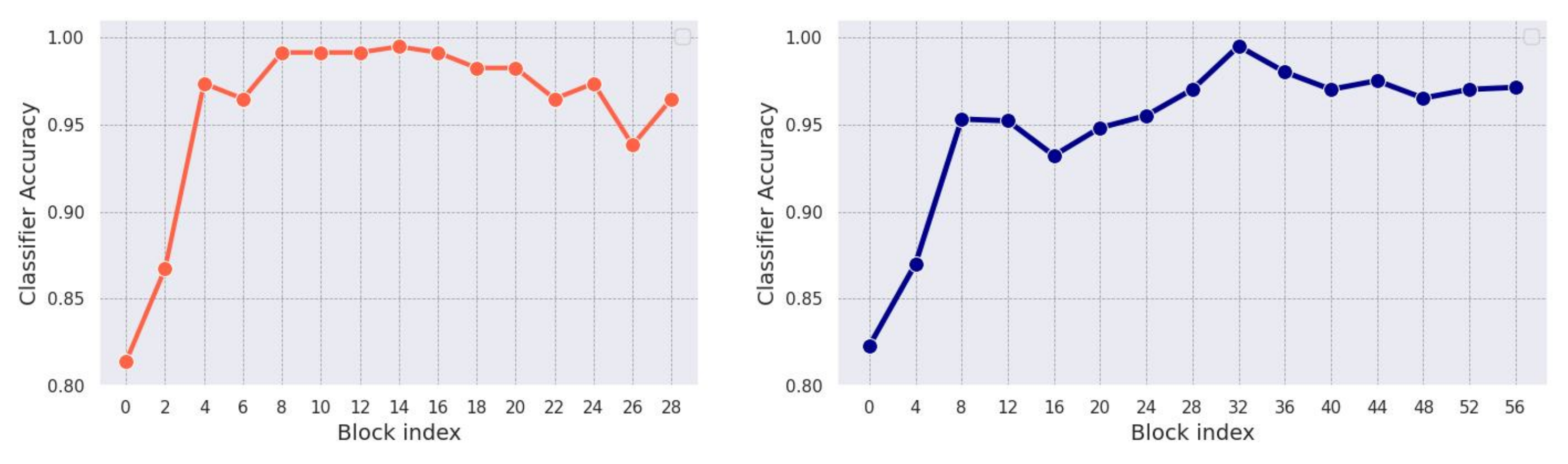}
    \caption{\textbf{Linear probing results on CogVideoX-5B~\cite{yang2024cogvideox} and Hunyuan Video~\cite{kong2024hunyuanvideo}.} We regard the layers that precede the higher classification accuracy as the layers of the diffusion encoder.}
    \label{fig:linear_probing}
\end{figure*}

\vspace{-0.2cm}
\textbf{Linear probing for locating denoising encoder} \ \ We first locate the encoder of the DAEs, as it is important to distill through the intermediate hidden state of the encoder for regularization. Similar to~\cite{xiang2023denoising, yu2024representation}, we conduct a linear probing experiment, where we train shallow linear classifiers for each hidden state from different layers to predict the class label of the noisy input. The intuition behind this experiment is that the layer with the highest classification accuracy retrieves the most interpretable hidden state, meaning that the layer yields the output of the encoder. To construct a dataset for the experiments, we use all the fine-tuning datasets we use for this paper by regarding each dataset as a class. We report the linear probing result for CogVideoX-5B and Hunyuan video in Fig.~\ref{fig:linear_probing}. Based on the result, we regard layers behind $14$th and $32$th as diffusion encoder for CogVideoX-5B and Hunyuan Video, respectively.

% Specifically, we extract hidden representations from different layers and train a lightweight linear classifier to predict the labels of the target dataset.
% We then measure the classification accuracy to evaluate the quality of the hidden representations at each layer.
% The results, shown in Fig.~\ref{fig:linear_probing_results}, reveal that certain intermediate layers exhibit stronger semantic representations.

% Based on the probing results, we select candidate layers around the peaks of classification performance for each model.
% For CogVideoX-5B (Fig.~\ref{fig:apd_layer_cog}) and HunyuanVideo (Fig.~\ref{fig:apd_layer_hun}), we qualitatively evaluate fine-tuned outputs using the candidate layers.
% From these comparisons, we select the final target layer that achieves the best balance between maintaining pretrained semantics and improving generation quality.
% These selected layers are then used when applying CREPA or REPA* for each respective model.
\vspace{-0.1cm}

\vspace{-0.2cm}
\textbf{Finding optimal hidden state layers for REPA*.} \ \ Having identified the diffusion encoder for each VDM, we next conduct experiments to determine the optimal intermediate layer of diffusion encoder for hidden state regularization. Specifically, we use VBench~\cite{huang2024vbench} to evaluate which layer yields the best performance on models regularized based on REPA*.

\begin{table*}[h]
\centering
\resizebox{\textwidth}{!}{
\begin{tabular}{l|ccccc}
\toprule
\textbf{Block index}(REPA*)  & Aesthetic Quality $\uparrow$  & Background Consistency $\uparrow$ & Motion Smoothness $\uparrow$  & Subject Consistency $\uparrow$ & Imaging Quality $\uparrow$ \\ 
\midrule
\textbf{8 }                      & 0.5640             & 0.9445             & 0.9874    & 0.9087     & 0.6635                                     \\ 
\textbf{10} (\textit{selected})                        & 0.5640             &\textbf{ 0.9459}\cellcolor[HTML]{DAE8FC}             & \textbf{0.9883\cellcolor[HTML]{DAE8FC}}    & \textbf{0.9143}\cellcolor[HTML]{DAE8FC}     & \textbf{0.6648}\cellcolor[HTML]{DAE8FC}                            \\ 
\textbf{12}         & \textbf{0.5645}\cellcolor[HTML]{DAE8FC}    & 0.9446    & 0.9754 & 0.9137 & 0.6640                       \\ 
\textbf{14}                         & 0.5643            & 0.9440            & 0.9765    & 0.9134     & 0.6547  \\

\midrule
\textbf{6} & 0.5460   & 0.9488  & 0.9857   &  0.8963   & 0.6245   \\ 
\textbf{8} (\textit{selected})      & \textbf{0.5519} \cellcolor[HTML]{DAE8FC} & \textbf{0.9497} \cellcolor[HTML]{DAE8FC}  & 0.9854     &   \textbf{0.9052} \cellcolor[HTML]{DAE8FC}  & 0.6258  \\ 
\textbf{10}   & 0.5478     & 0.9490    &  \textbf{0.9861} \cellcolor[HTML]{DAE8FC}  & 0.9002 & \textbf{0.6265} \cellcolor[HTML]{DAE8FC}        \\ 
\textbf{12} & 0.5479   & 0.9457  & 0.9847   &   0.8994   & 0.6243   \\ 
\bottomrule
\end{tabular}
}
\caption{\textbf{VBench~\cite{huang2024vbench} results on models trained with REPA under different hidden state layers}. We experimented over Hunyuan Video~\cite{kong2024hunyuanvideo} (top) and CogVideoX-5B~\cite{yang2024cogvideox} (bottom). Selected indices are marked as (\textit{selected}); best scores are highlighted.}

% \caption{\textbf{VBench~\cite{huang2024vbench} evaluation on Hunyuan Video~\cite{kong2024hunyuanvideo} (top) and CogVideoX-5B~\cite{yang2024cogvideox} (bottom) for selecting REPA block index.} 
% We report scores across different block indices, with the selected one marked as (\textit{selected}) and the best score per metric highlighted.}

% \caption{\textbf{VBench~\cite{huang2024vbench} evaluation for identifying optimal hidden state block in REPA.} 
% We report video-level VBench scores across different block indices of REPA when applied to Hunyuan Video~\cite{kong2024hunyuanvideo} (top) and CogVideoX-5B~\cite{yang2024cogvideox} (bottom). 
% Each row corresponds to a different block index, and the selected index used in our final experiments is marked as (\textit{selected}). 
% The highest score per metric is highlighted.}

\label{tab:vbench_hunyuan}
\end{table*}
\vspace{-0.2cm}
\section{Implementation details for empirical analysis via CKNNA}
\label{app:analysis}
\vspace{-0.2cm}
We used Disney~\cite{wildheart2024disney} dataset for fine-tuning CogVideoX-5B~\cite{yang2024cogvideox} using DINOv2~\cite{oquabdinov2} as a pretrained encoder $E$ for the distillation-based regularization. Following REPA~\cite{yu2024representation}, we evaluate with hidden states for timesteps $0 \leq t \leq 0.5$. For CKA~\cite{kornblith2019similarity}, we follow the exact formulation from REPA. Also, we use inner product as a kernel, and use $k=10$ nearest neighbors for CKNNA~\cite{huh2024platonic}. When visualizing the results using box plot, each data point is calculated per frame. For instance, a video whose number of frames in VAE latent space is $49$ yields $48$ measurements if it is a cross-frame measurement or $49$ measurements if it is a measurement for each frame. We remove the lower and upper $3\%$ of the measurements.

\vspace{-0.2cm}
\section{More qualitative results}
\label{app:more}
\vspace{-0.2cm}
Here, we provide additional qualitative results. \textit{We strongly recommend to refer to the project page for accurate qualitative comparison}.

\newpage

%%%%% Crush %%%%%

\begin{figure}
    \centering
    \includegraphics[width=\textwidth]{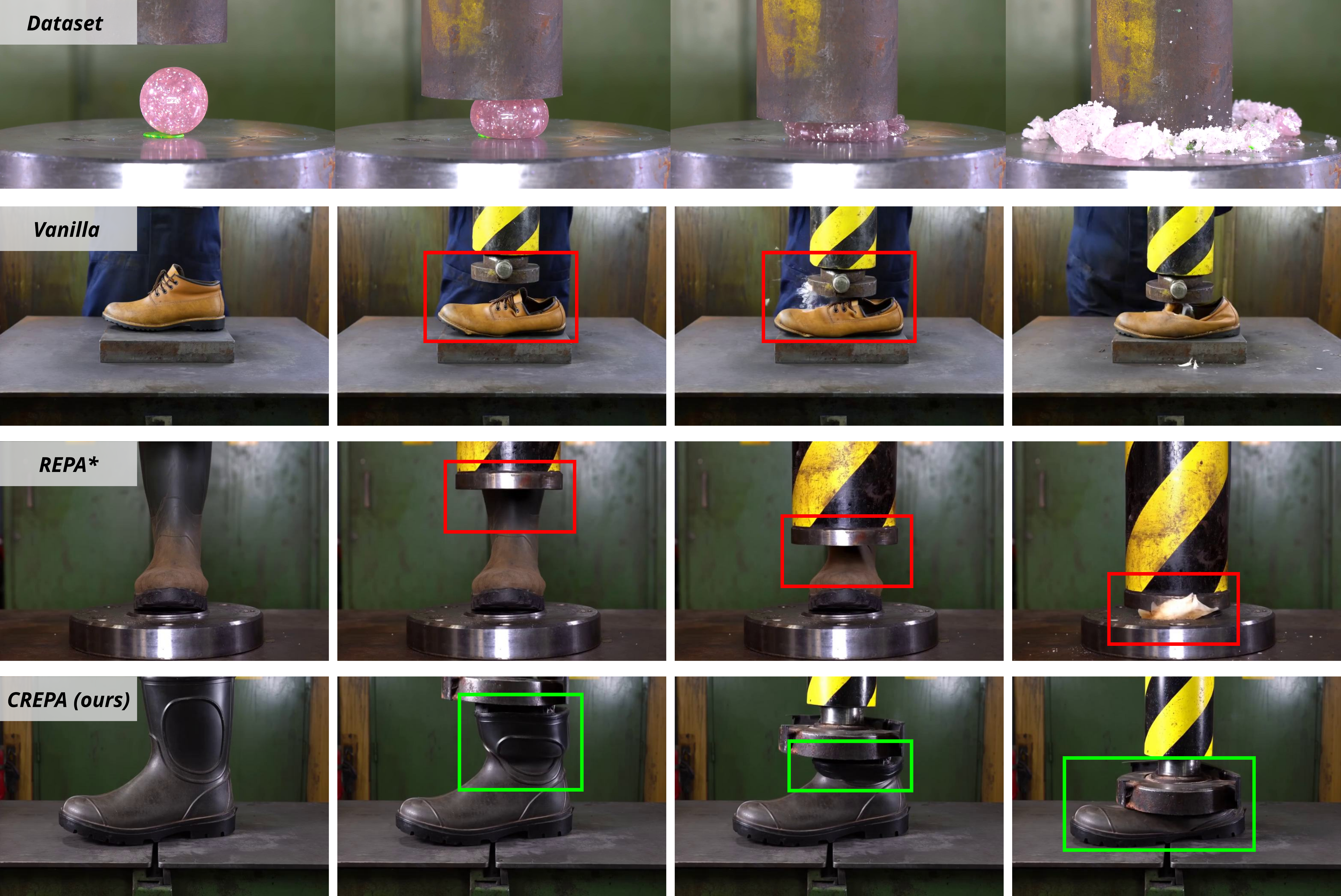}
    \caption{\textbf{Videos generated by Hunyuan Video~\cite{kong2024hunyuanvideo} fine-tuned on Crush~\cite{finetrainers_crush_smol} dataset.}}
    \label{fig:apd_crush_hun}
\end{figure}
%A rubber boot is placed on the platform. As the hydraulic press moves down, the boot compresses and wrinkles before bursting at the seams.

%%%%% Crush %%%%%

\begin{figure}
    \centering
    \includegraphics[width=\textwidth]{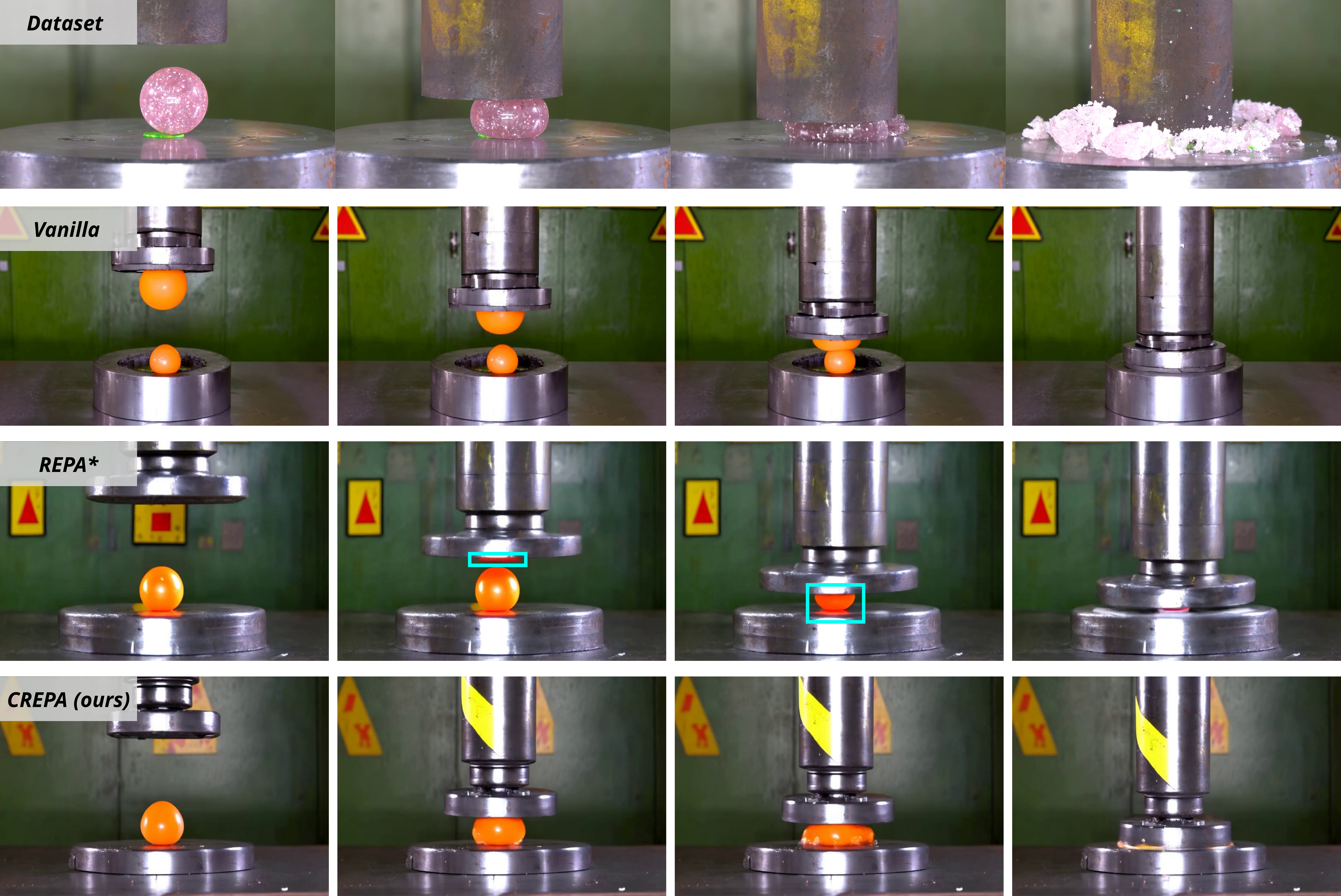}
    \caption{\textbf{Videos generated by CogVideoX-5B~\cite{yang2024cogvideox} fine-tuned on Crush~\cite{finetrainers_crush_smol} dataset.}}
    \label{fig:apd_crush}
\end{figure}
%A large, cylindrical object is seen pressing down on a small orange ball, causing it to flatten as if it were under a hydraulic press. The background features a green wall with yellow and red warning signs.

%%%%% Cakeify %%%%%

\begin{figure}
    \centering
    \includegraphics[width=\textwidth]{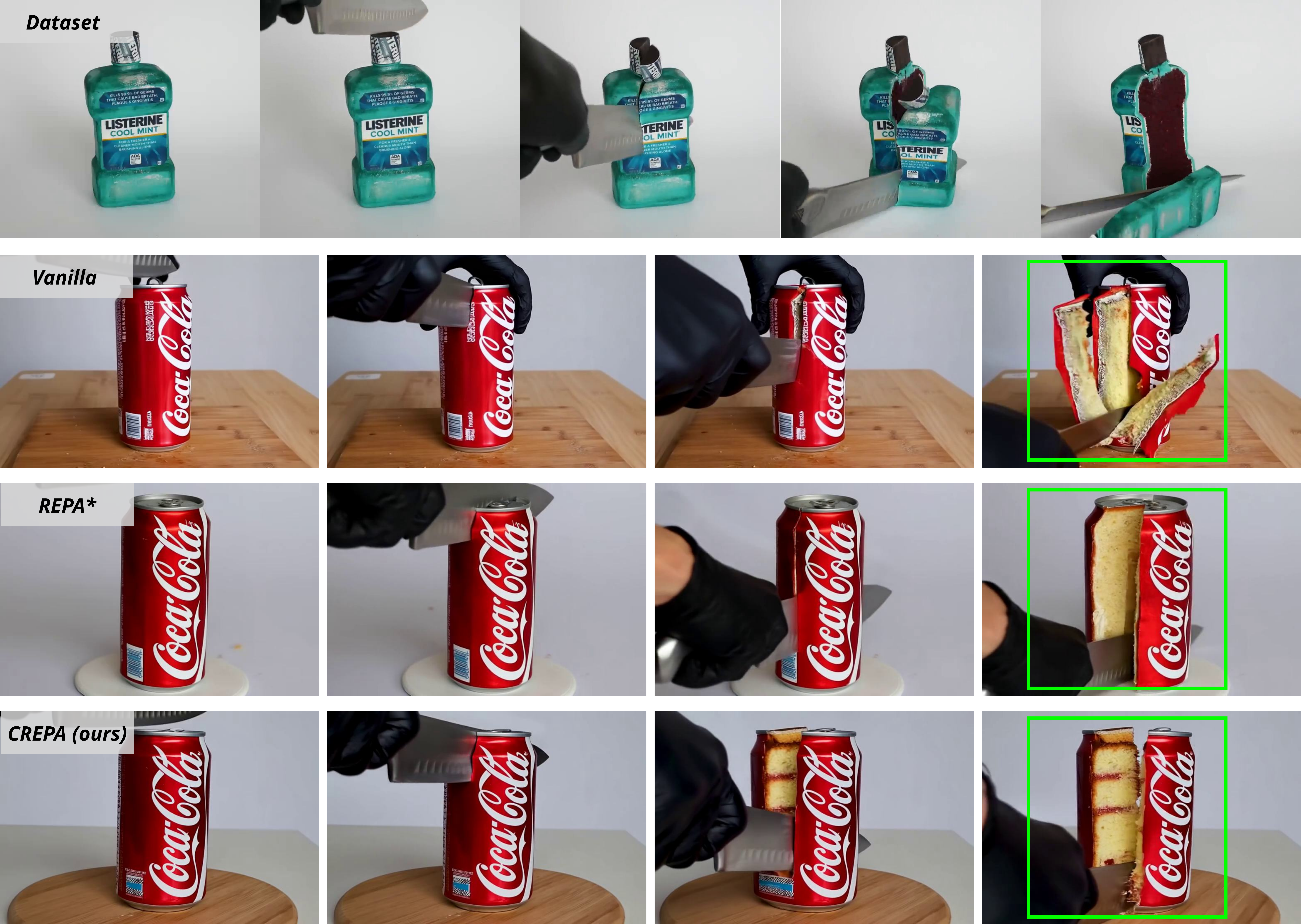}
    \caption{\textbf{Videos generated by Hunyuan Video~\cite{kong2024hunyuanvideo} fine-tuned on Cakeify~\cite{finetrainers_cakeify_smol} dataset.}}
    \label{fig:apd_cake_hun}
\end{figure}
%A hand wearing a black glove holds a knife, slicing through a Coca-Cola can that has been transformed into a hyper-realistic prop cake. The cake is cut in half, revealing its cake-like interior.

\begin{figure}
    \centering
    \includegraphics[width=\textwidth]{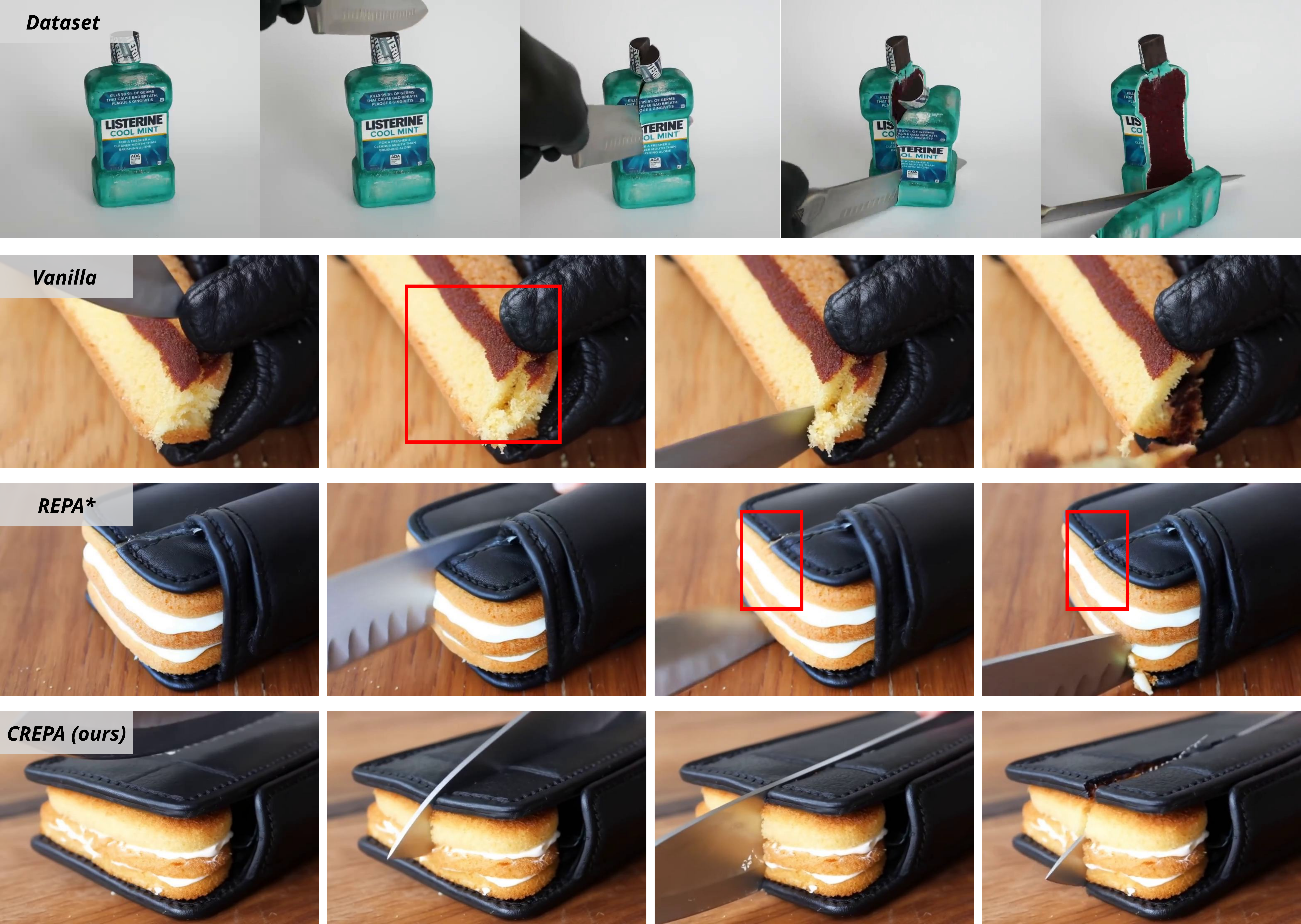}
    \caption{\textbf{Videos generated by Hunyuan Video~\cite{kong2024hunyuanvideo} fine-tuned on Cakeify~\cite{finetrainers_cakeify_smol} dataset.}}
    \label{fig:apd_cake_hun2}
\end{figure}
%A black leather wallet rests on a wooden surface. A sharp blade cuts into the material, exposing layers of sponge and frosting beneath the realistic edible leather texture.

\begin{figure}
    \centering
    \includegraphics[width=\textwidth]{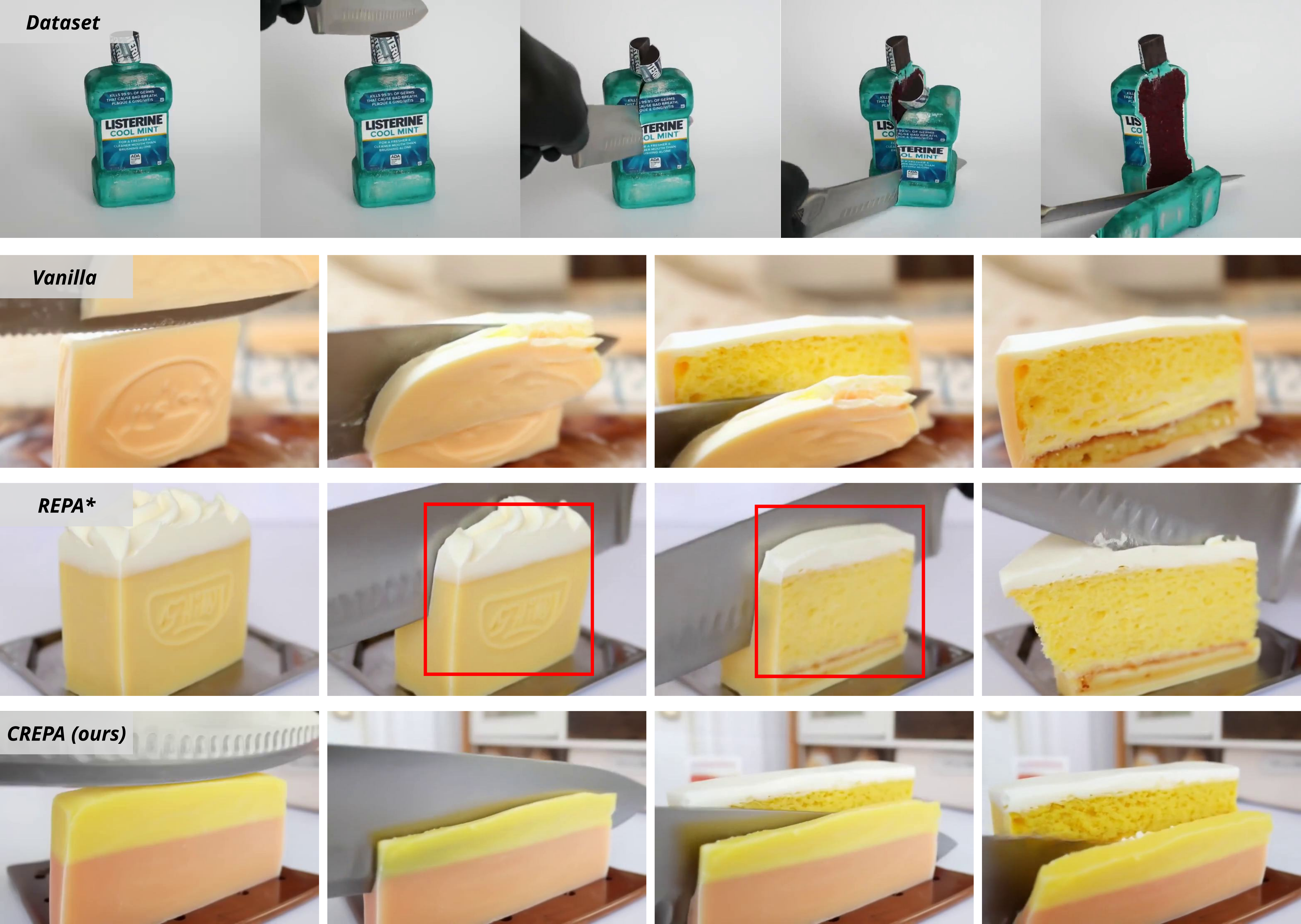}
    \caption{\textbf{Videos generated by Hunyuan Video~\cite{kong2024hunyuanvideo} fine-tuned on Cakeify~\cite{finetrainers_cakeify_smol} dataset.}}
    \label{fig:apd_cake_hun3}
\end{figure}
%A bar of soap sits in a soap dish, its pastel color catching the light. A blade smoothly cuts through the bar, revealing layers of lemon cake and frosting beneath the glossy icing.

\begin{figure}
    \centering
    \includegraphics[width=\textwidth]{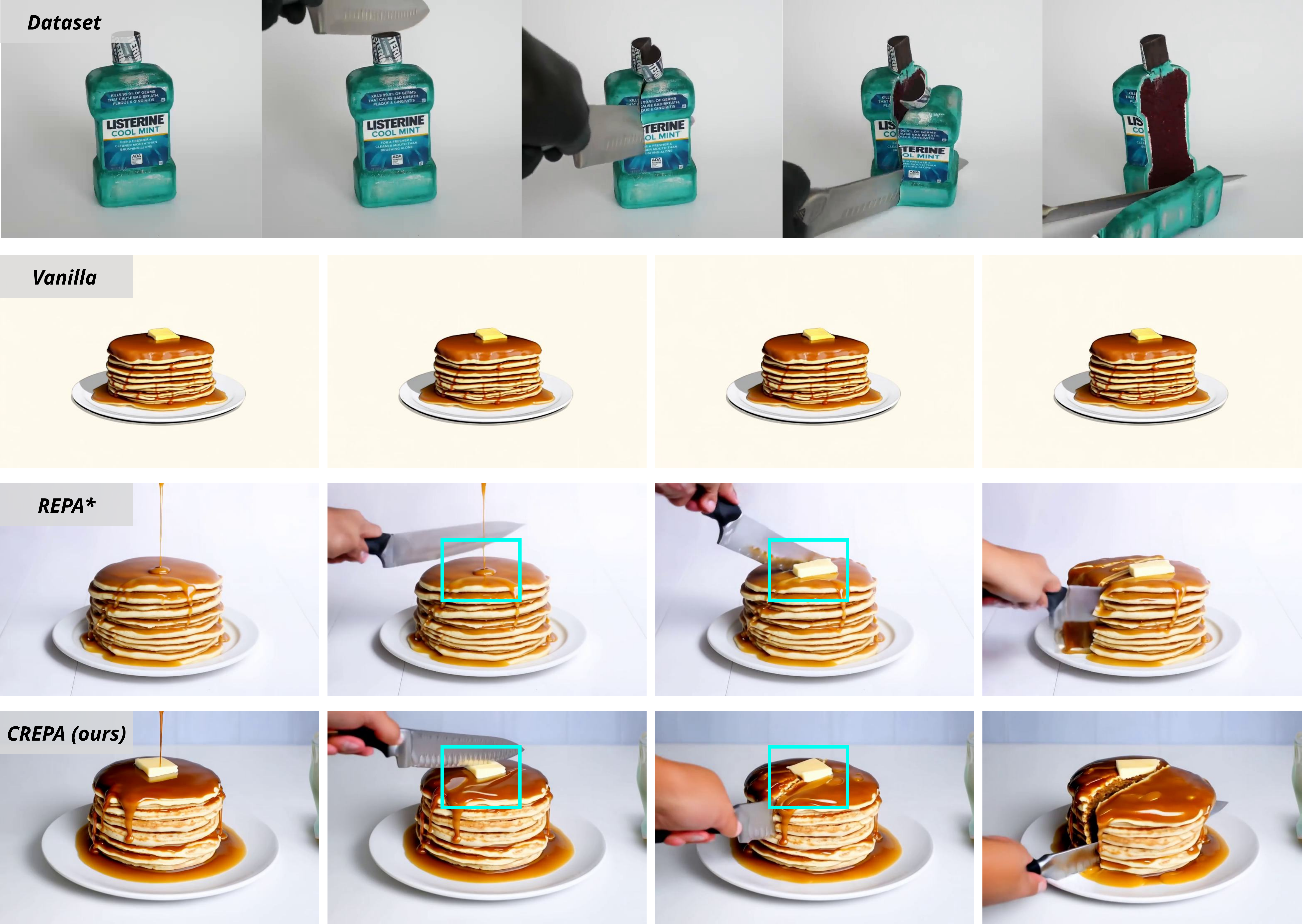}
    \caption{\textbf{Videos generated by CogVideoX-5B~\cite{yang2024cogvideox} fine-tuned on Cakeify~\cite{finetrainers_cakeify_smol} dataset.}}
    \label{fig:apd_cake1}
\end{figure}
%A stack of pancakes sits on a white plate, drizzled with syrup and butter. A hand with a knife slices through the stack, unveiling that the entire dish is actually a cake, complete with pancake-textured fondant and caramel-flavored layers.

\begin{figure}
    \centering
    \includegraphics[width=\textwidth]{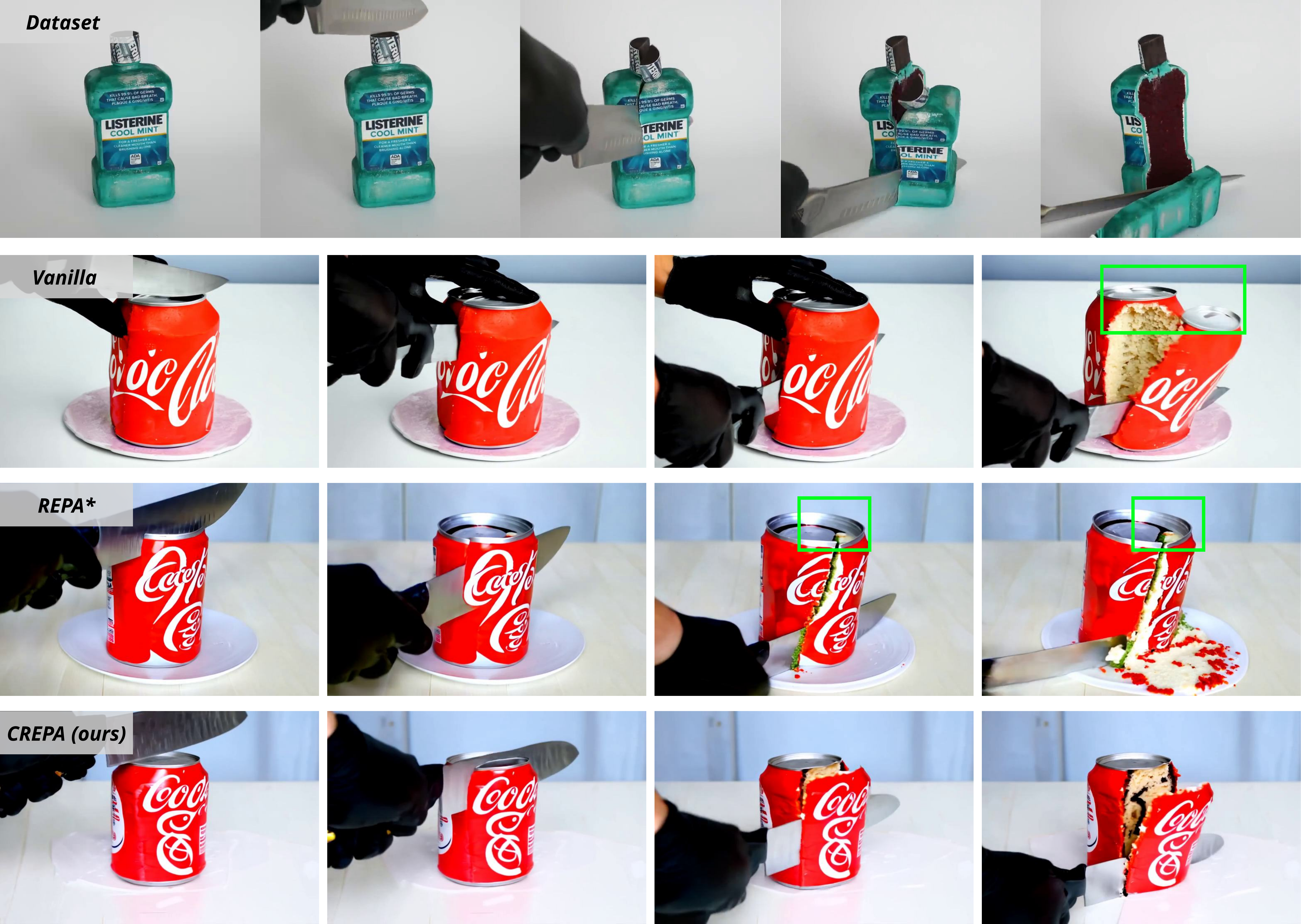}
    \caption{\textbf{Videos generated by CogVideoX-5B~\cite{yang2024cogvideox} fine-tuned on Cakeify~\cite{finetrainers_cakeify_smol} dataset.}}
    \label{fig:apd_cake3}
\end{figure}
%A hand wearing a black glove holds a knife, slicing through a Coca-Cola can that has been transformed into a hyper-realistic prop cake. The cake is cut in half, revealing its cake-like interior.

%%%%% Squish %%%%%

\begin{figure}[t!]
    \centering
    \includegraphics[width=\textwidth]{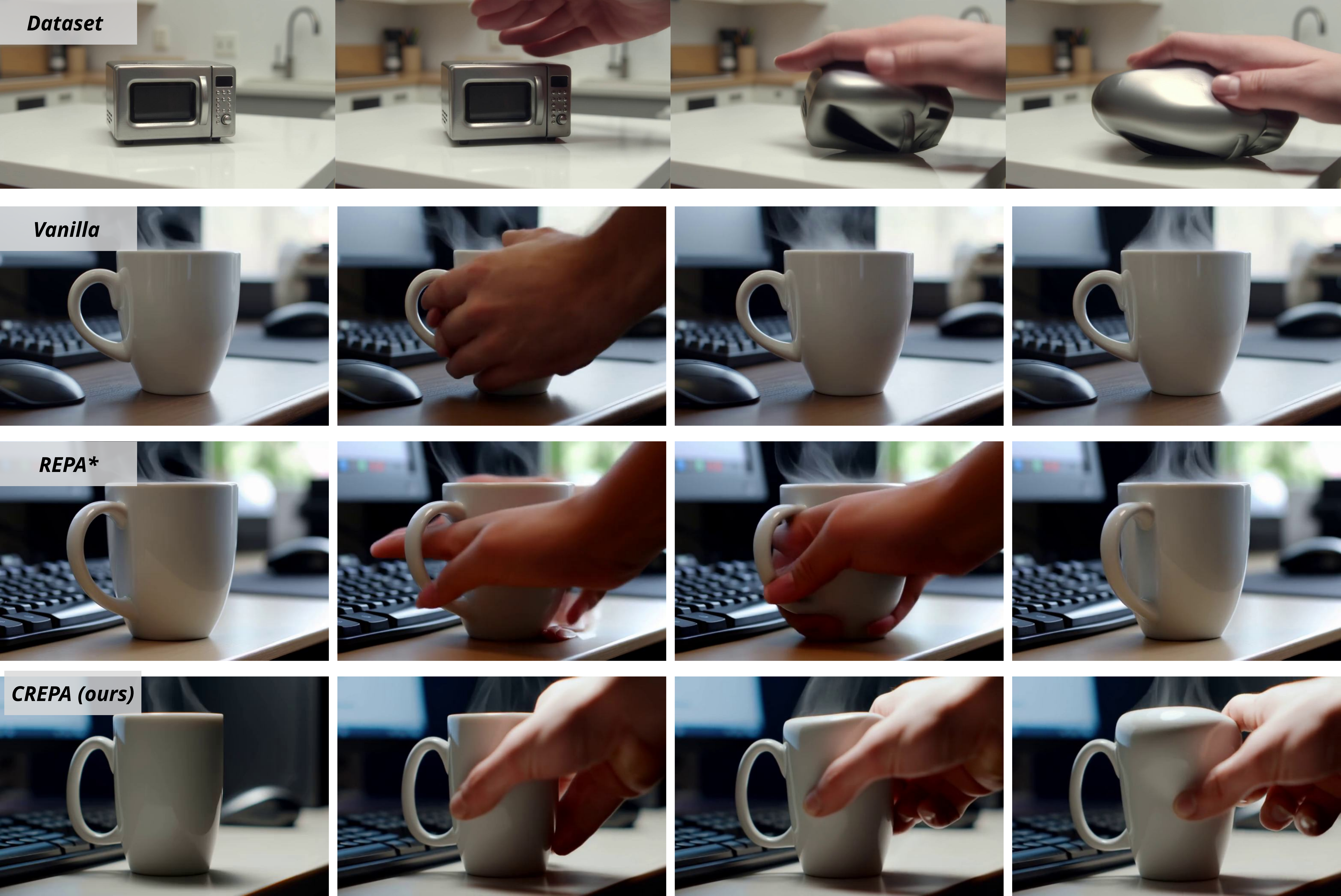}
    \caption{\textbf{Videos generated by Hunyuan Video~\cite{kong2024hunyuanvideo} fine-tuned on Squish~\cite{finetrainers_squish_pika} dataset.}}
    \label{fig:apd_squish_hun2}
\end{figure}
%A steaming cup of coffee sits on a desk. A hand gently presses down on the ceramic mug, which bends and flattens like a stress ball, losing all its rigid form.

\begin{figure}[t!]
    \centering
    \includegraphics[width=\textwidth]{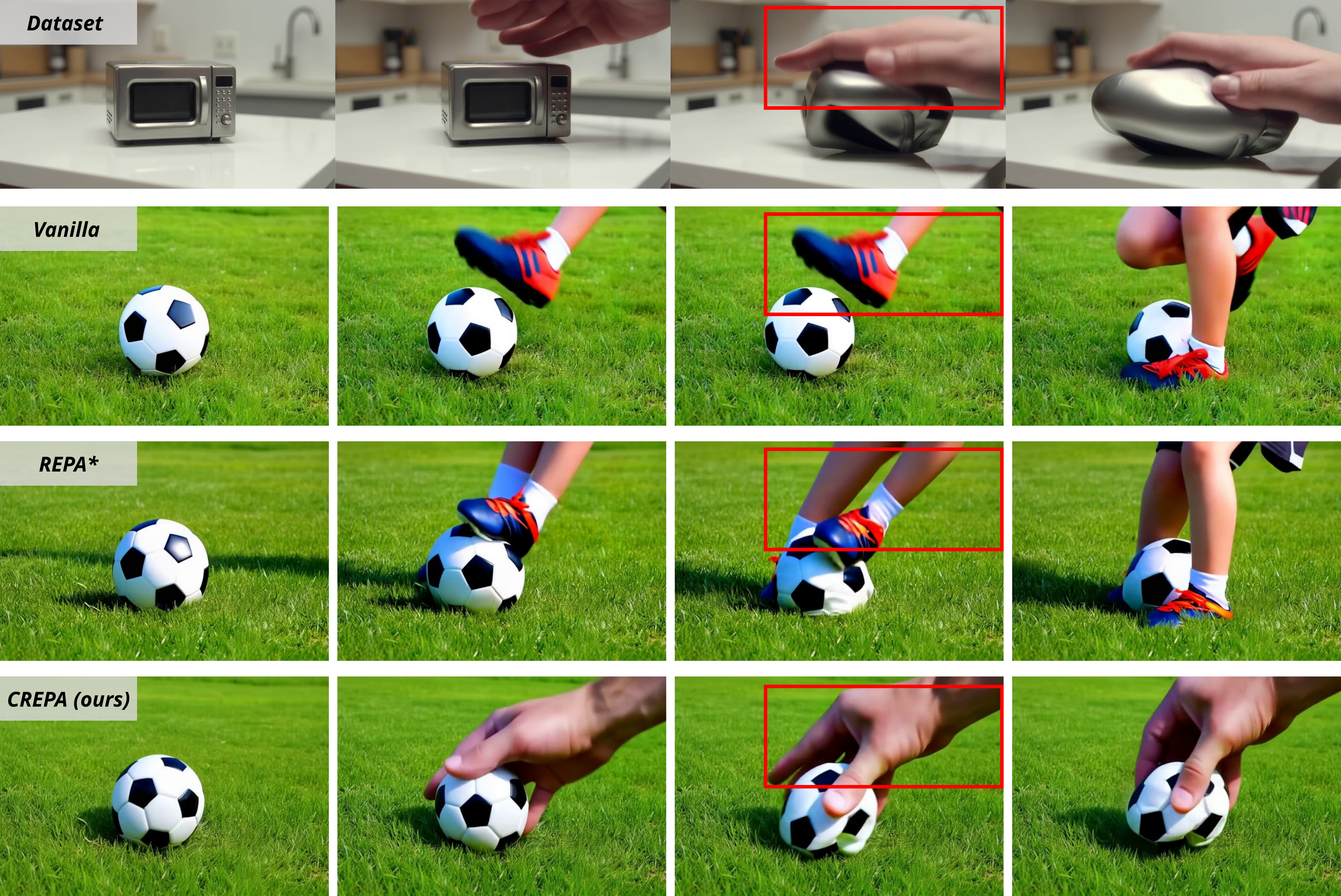}
    \caption{\textbf{Videos generated by Hunyuan Video~\cite{kong2024hunyuanvideo} fine-tuned on Squish~\cite{finetrainers_squish_pika} dataset.}}
    \label{fig:apd_squish_hun}
\end{figure}
%A soccer ball is placed on a grassy field. A child’s foot kicks it, but instead of bouncing away, it deforms completely and remains squashed on the ground.

\begin{figure}[t!]
    \centering
    \includegraphics[width=\textwidth]{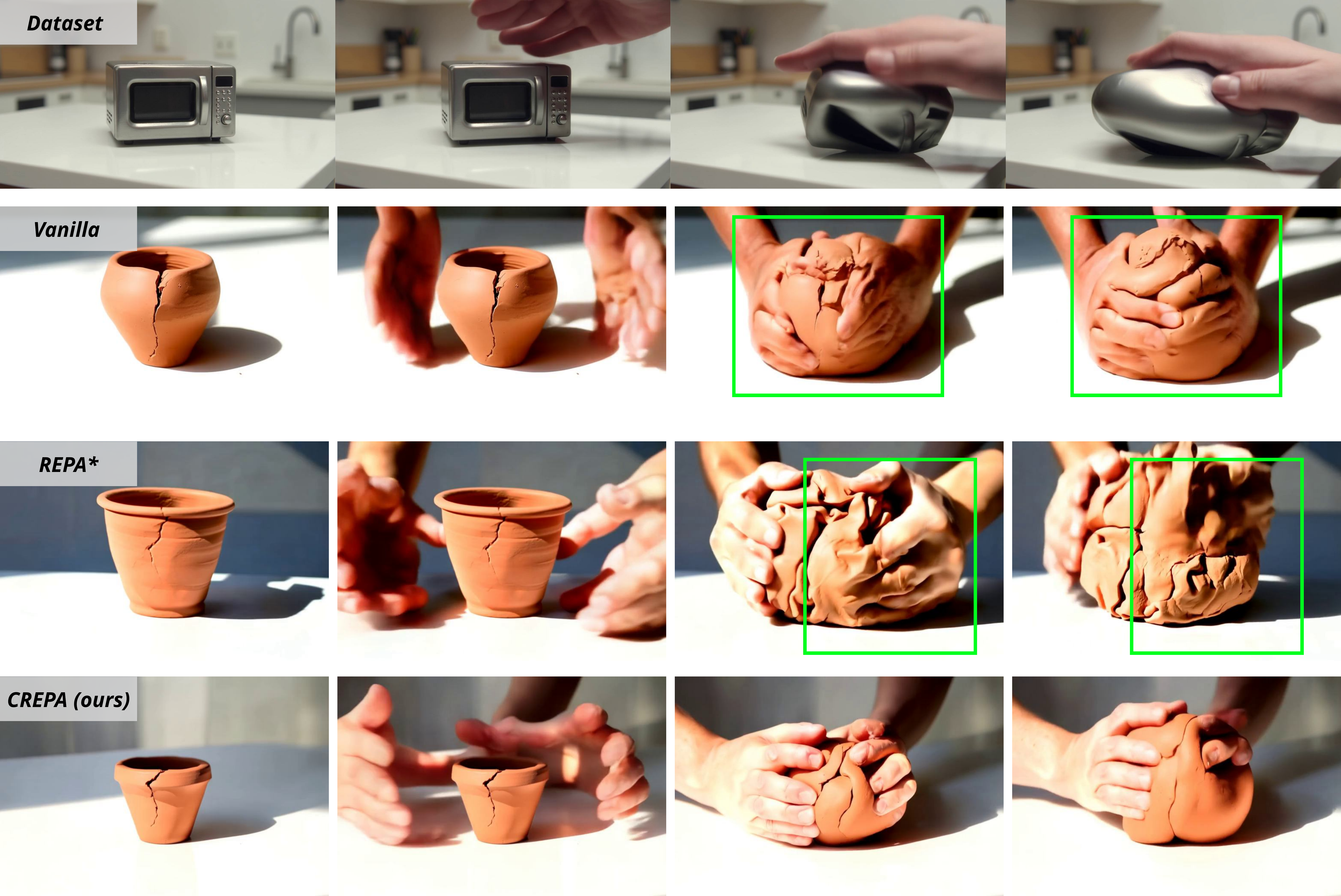}
    \caption{\textbf{Videos generated by CogVideoX-5B~\cite{yang2024cogvideox} fine-tuned on Squish~\cite{finetrainers_squish_pika} dataset.}}
    \label{fig:apd_squish2}
\end{figure}
%A terracotta pot with a visible crack sits centered on a white surface, bathed in sunlight. Two hands enter the frame, positioning themselves around the pot. The hands then begin to press and mold the pot, the clay beginning to rise from the opening. The clay is reshaped and compressed, transforming the pot into a bulbous, amorphous lump. The final shot displays the morphed clay shape standing upright on the white surface.

%%%%% Disney %%%%%
\begin{figure}
    \centering
    \includegraphics[width=\textwidth]{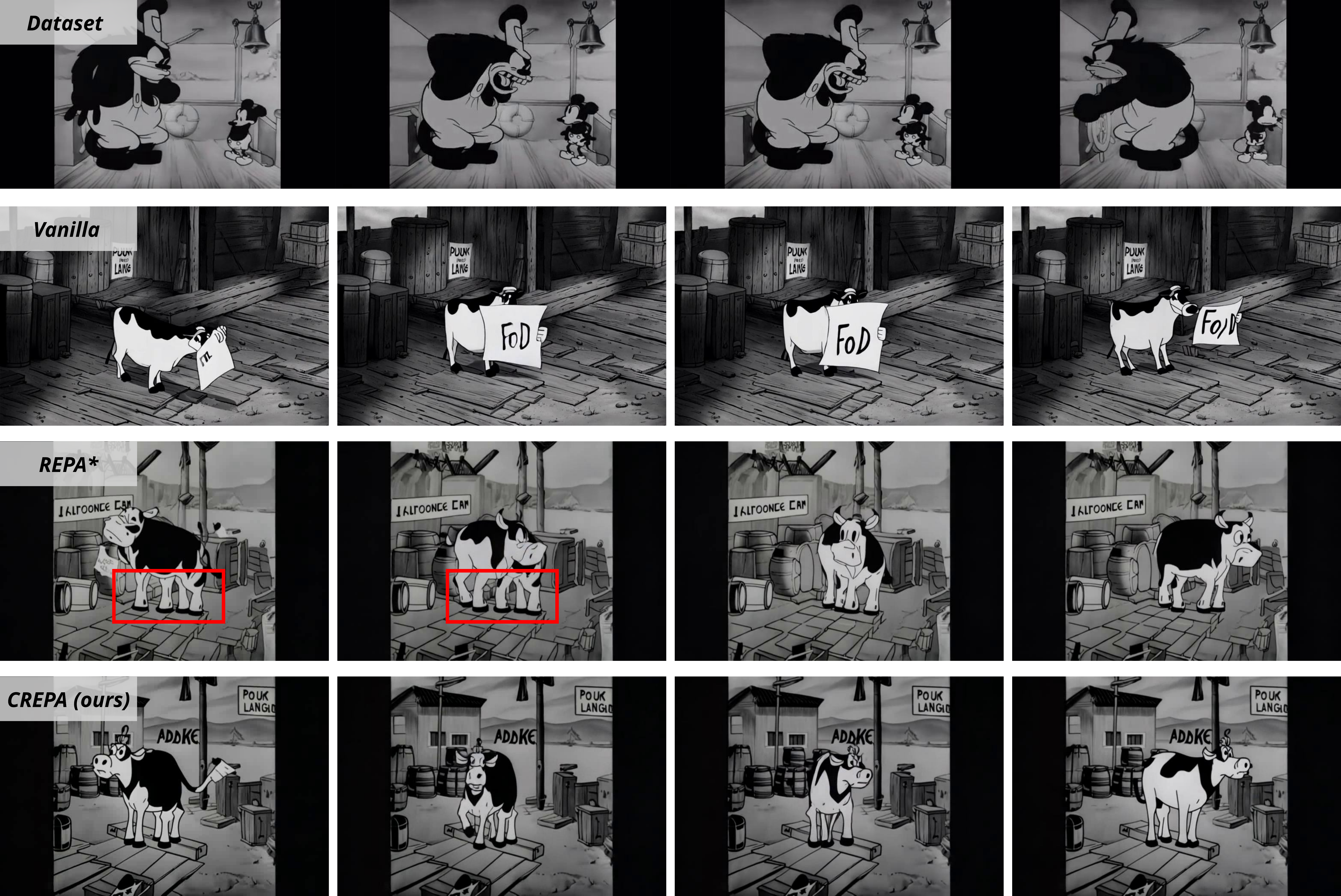}
    \caption{\textbf{Videos generated by CogVideoX-5B~\cite{yang2024cogvideox} fine-tuned on Disney~\cite{wildheart2024disney} dataset.}}
    \label{fig:apd_disney1}
\end{figure}
%A black-and-white animated scene unfolds on a semi-rural dock, with a cow standing on wooden planks, holding a piece of paper with 'FOB' written on it. The cow is the central focus, amidst static barrels, crates, and a 'PODUNK LANDING' sign in the background. The atmosphere remains calm and still, with the cow's presence subtly shifting the narrative's tone. A sign of pause or anticipation, the scene is frozen in time, inviting the viewer to ponder the story's next development.

\begin{figure}
    \centering
    \includegraphics[width=\textwidth]{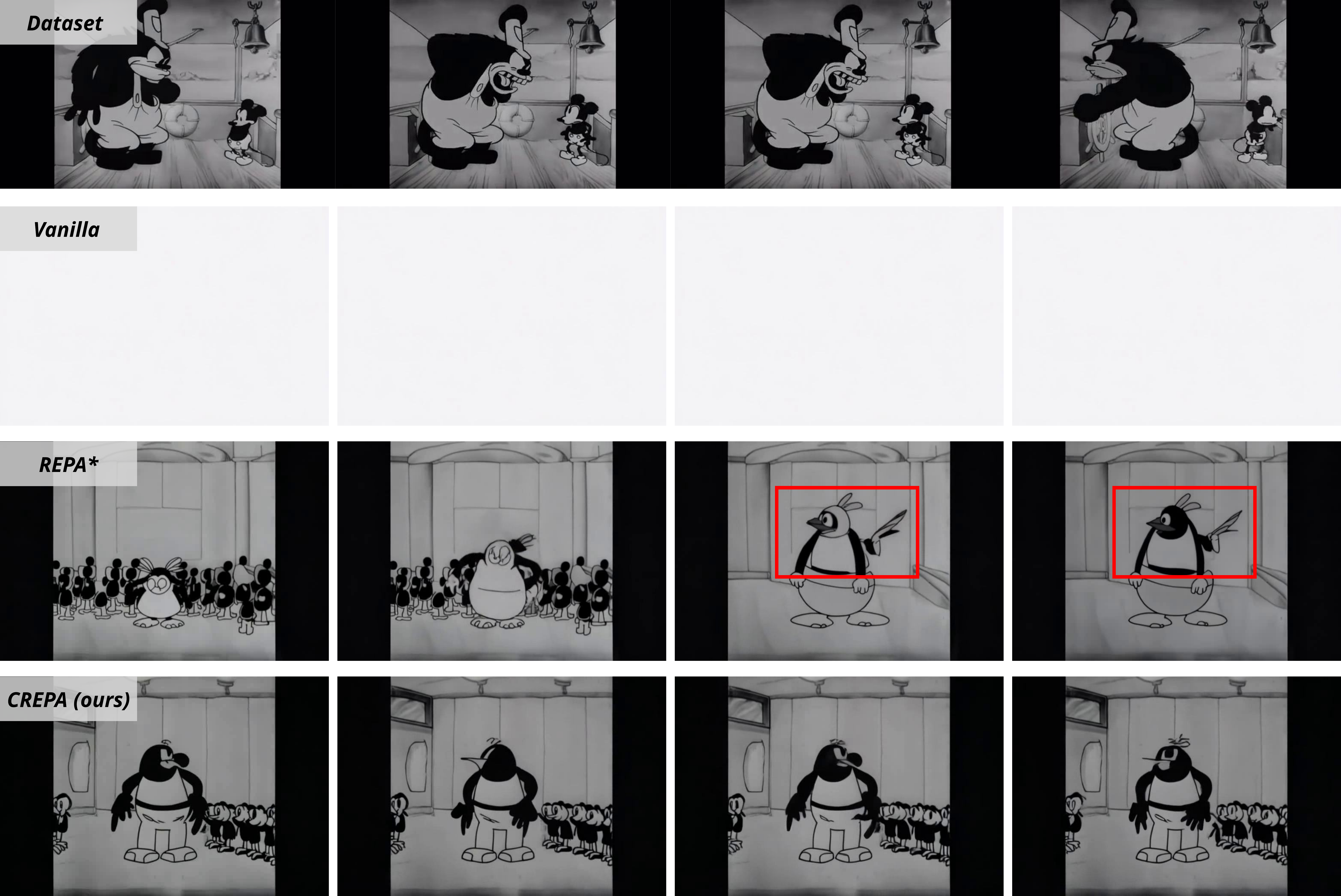}
    \caption{\textbf{Videos generated by CogVideoX-5B~\cite{yang2024cogvideox} fine-tuned on Disney~\cite{wildheart2024disney} dataset.}}
    \label{fig:apd_disney}
\end{figure}
%A black-and-white animated video showcases a central character with a round body and large ears standing in an indoor setting with a plain background. The character is surrounded by smaller figures, displaying various expressions of interest or curiosity. As the video progresses, subtle changes occur among the figures, suggesting movement and reactions. The scene transitions to focus on a single bird perched on a perch, with its posture and expression changing subtly throughout the frames, showing signs of activity.

%%%%% Tom and Jerry %%%%%

\begin{figure}[t!]
    \centering
    \includegraphics[width=\textwidth]{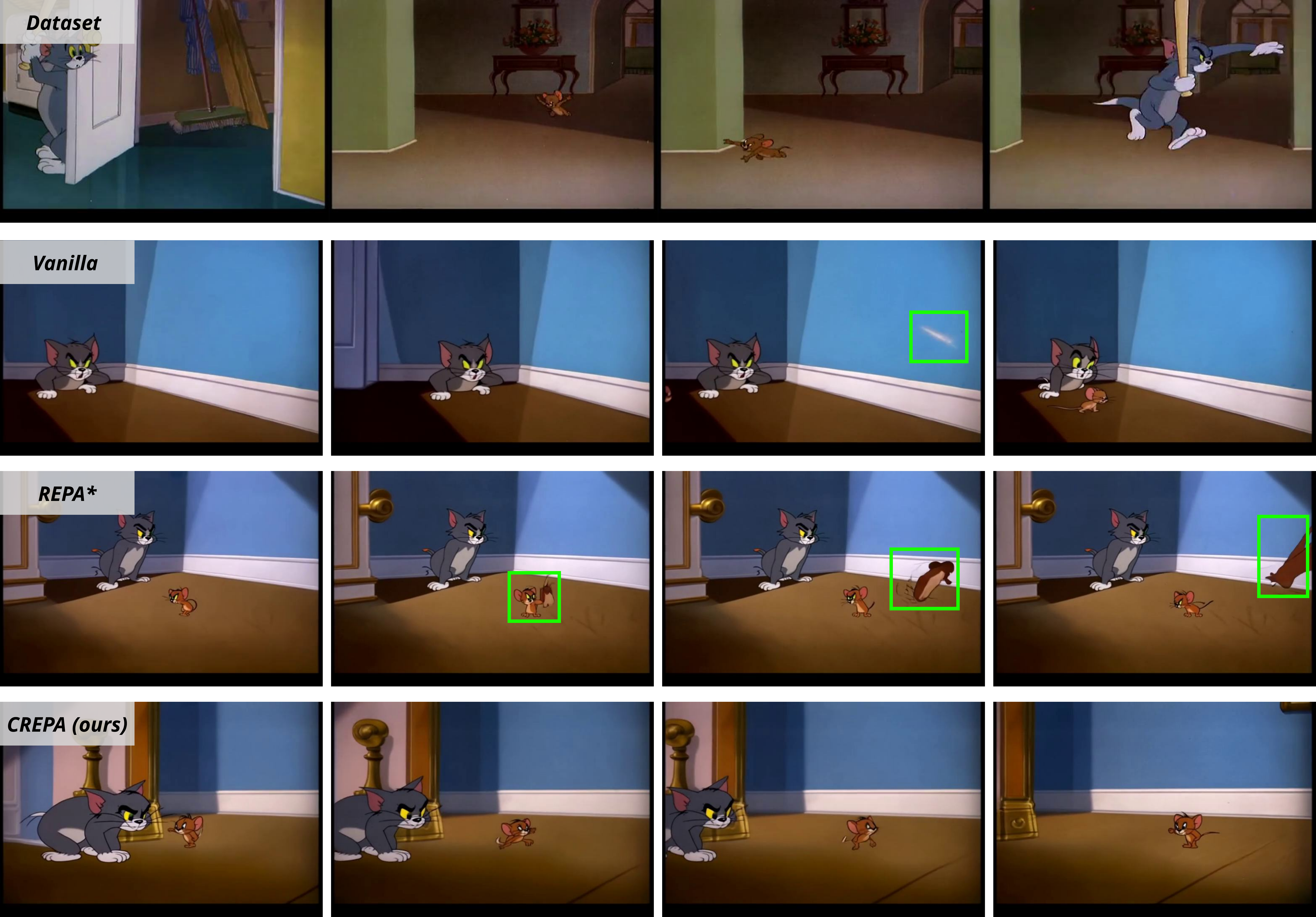}
    \caption{\textbf{Videos generated by CogVideoX-5B~\cite{yang2024cogvideox} fine-tuned on Tom and Jerry~\cite{wildheart2024tomjerry} dataset.}}
    \label{fig:apd_tom}
\end{figure}
%Tom, the mischievous cat, is crouched in the corner of a dimly lit room, his eyes fixed on Jerry, the clever mouse, who scurries across the wooden floor. The room is adorned with a blue wall and a golden doorknob, adding a touch of elegance to the otherwise mundane setting. Tom's fur is a mix of gray and white, blending seamlessly with the shadows cast by the flickering light. Jerry, on the other hand, is a vibrant mix of brown and white, his nimble movements creating a stark contrast against the stillness of the room. The tension between the two characters is palpable, as they engage in a silent battle of wit and agility. The scene is a classic representation of the timeless rivalry between Tom and Jerry, captured in a moment of suspense and anticipation.

% DL3DV

\begin{figure}
    \centering
    \includegraphics[width=\textwidth]{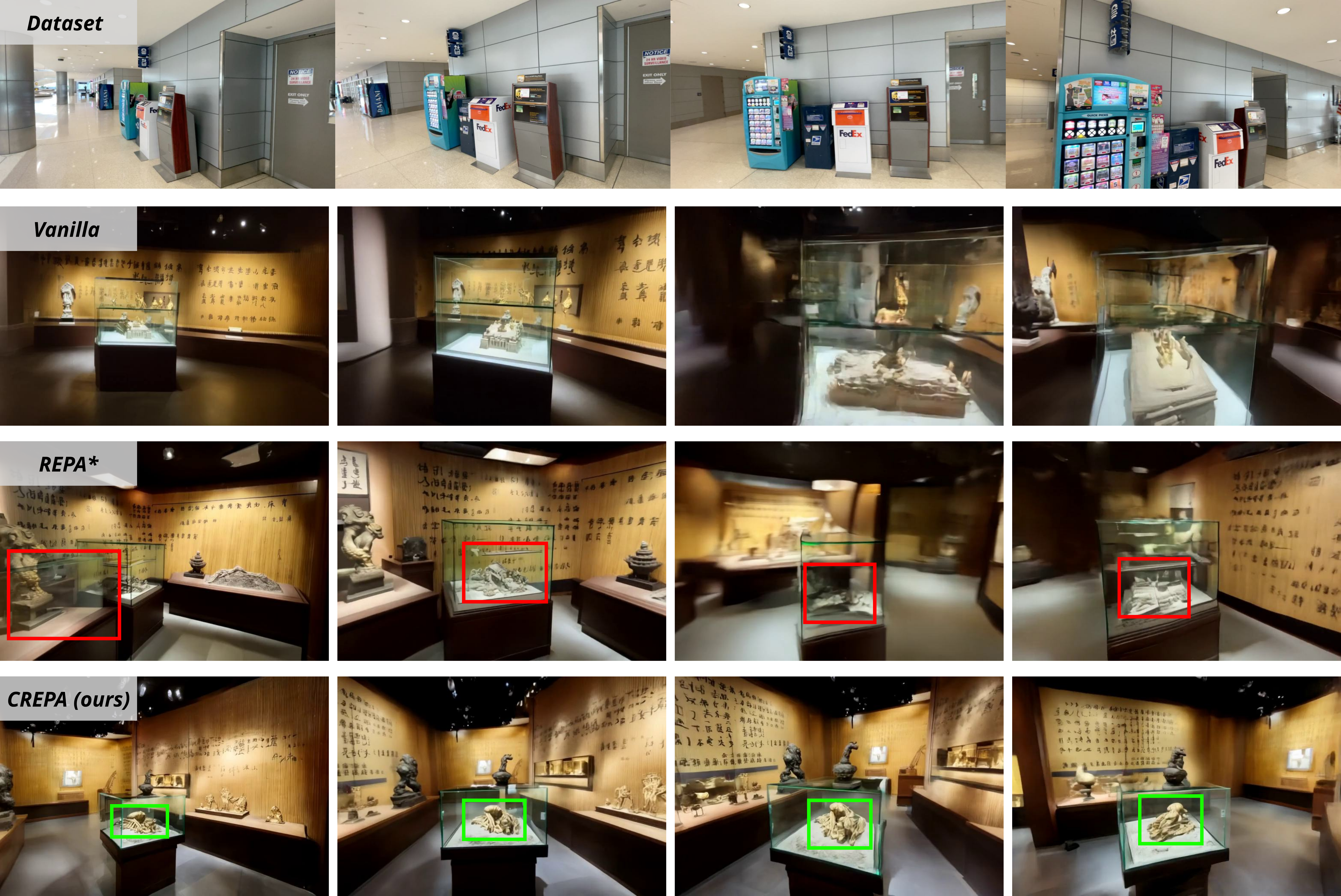}
    \caption{\textbf{Videos generated by CogVideoX-5B~\cite{yang2024cogvideox} fine-tuned on DL3DV~\cite{ling2024dl3dv} dataset.}}
    \label{fig:apd_dl3dv}
\end{figure}
%The video explores a traditional Chinese museum, starting with a serene interior featuring a glass display case with golden figurines and a bamboo wall with calligraphy. As the video continues, various scenes show the museum's historical artifacts, including a lion statue, a model of an ancient Chinese architectural structure, and a skeleton with animal bones. The exhibit also includes a display case with a human skeleton, a model of an ancient burial site, and a bronze incense burner. The museum's ambiance is tranquil, with soft lighting and a blend of historical and modern elements, culminating in a display of golden figurines and a model of an ancient Chinese architectural complex.

%%%%% Scenes %%%%%
\begin{figure}[t!]
    \centering
    \includegraphics[width=\textwidth]{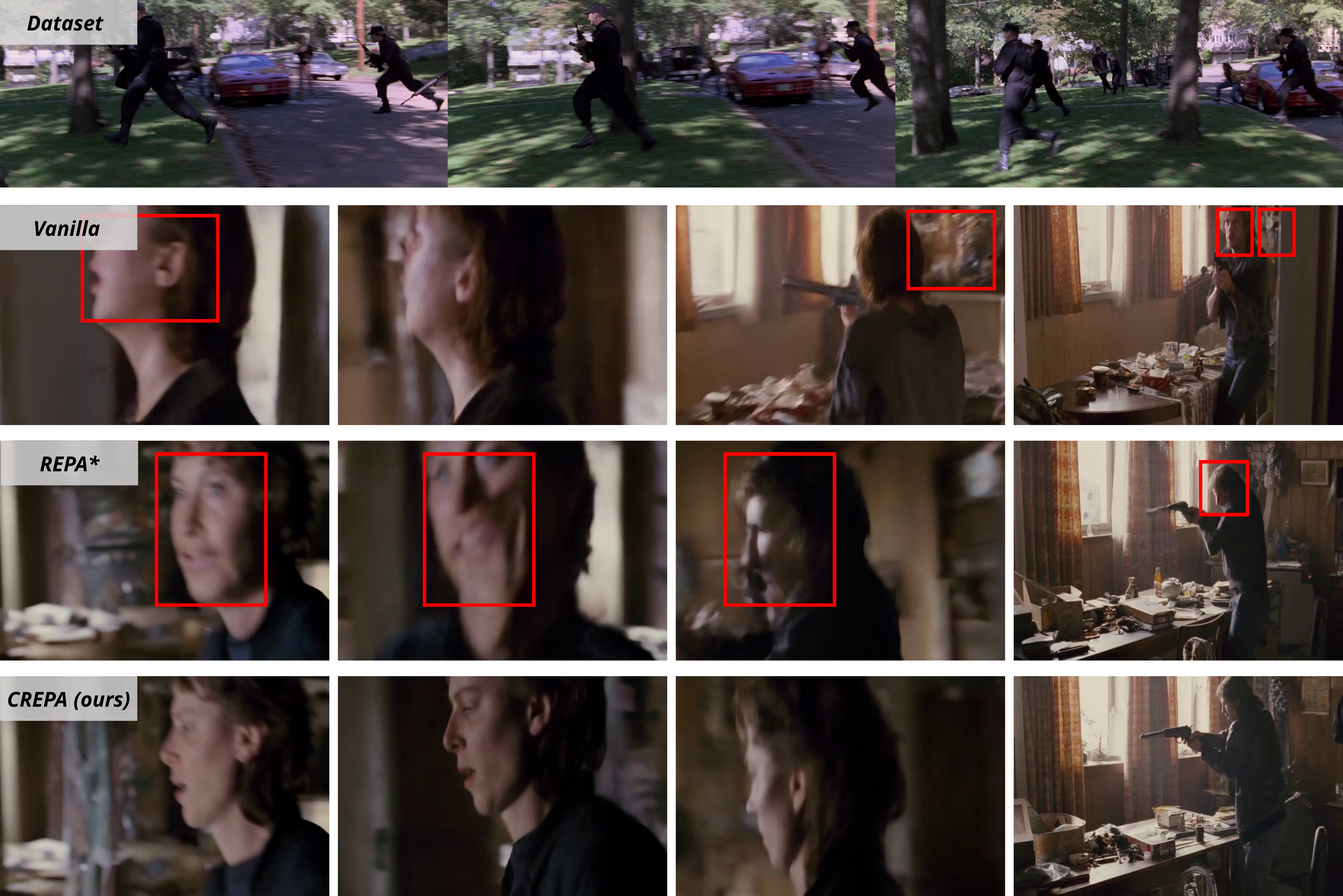}
    \caption{\textbf{Videos generated by Hunyuan Video~\cite{kong2024hunyuanvideo} fine-tuned on Scenes~\cite{finetrainers_cakeify_smol} dataset.}}
    \label{fig:apd_scenes_hun2}
\end{figure}
%A person is seen standing in a cluttered room, with a table covered in various items. The scene then shifts to a man holding a gun, seemingly in a threatening stance. The man's presence creates a sense of tension and danger in the room.

\begin{figure}[t!]
    \centering
    \includegraphics[width=\textwidth]{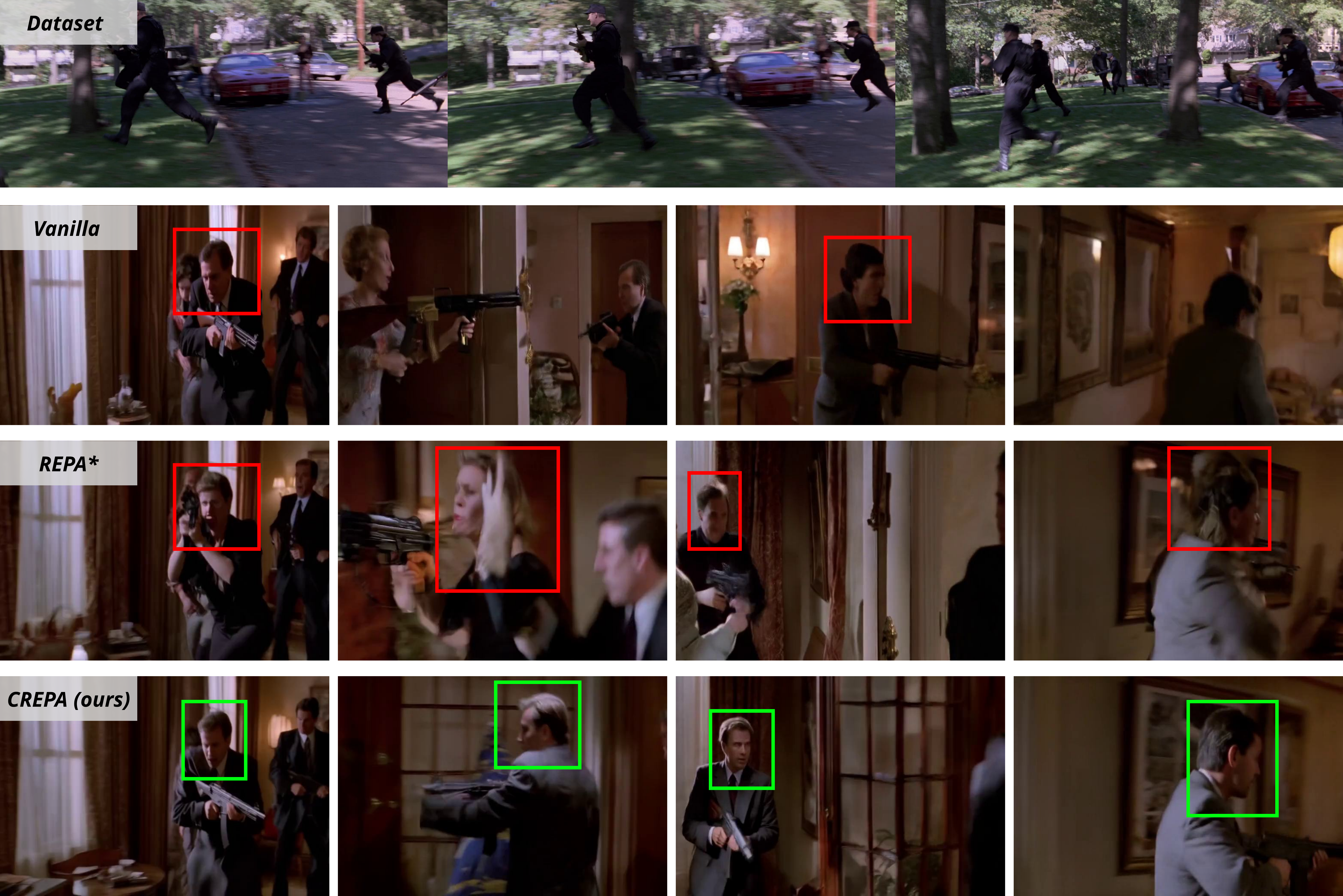}
    \caption{\textbf{Videos generated by Hunyuan Video~\cite{kong2024hunyuanvideo} fine-tuned on Scenes~\cite{bigdata-pw_scenes} dataset.}}
    \label{fig:apd_scenes_hun}
\end{figure}
%In the video, a group of individuals are seen entering a room, each holding a gun. The scene progresses as they move deeper into the room, flattening objects as if they were under a hydraulic press. The room is adorned with framed pictures on the walls, adding to the tension of the scene.

\clearpage

\section{Text prompts used for generation}

\label{app:textgen}
% \clearpage  % 현재까지 쌓인 float들(figure 등)을 모두 출력하고 새 페이지 시작
\phantomsection  % 하이퍼링크용 (선택)
\addcontentsline{toc}{section}{Appendix Prompts}  % 목차 항목 추가 (선택)

% Somewhere in the main body (not appendix)
\begin{tcolorbox}[
    title=Prompts for figures in main pages (Fig. 4–6),
    colback=pastelblue2,
    colframe=pastelblue,
    boxrule=0.4pt,
    arc=1mm,
    sharp corners=south,
    breakable,
    parskip=false,
    left=6pt, right=6pt, top=6pt, bottom=6pt
]
\small

\textbf{Fig. 4:}
\begin{itemize}[leftmargin=1em]
  \item A colorful puzzle ball is being crushed by a large metal cylinder, which flattens the objects as if they were under a hydraulic press.
\end{itemize}

\textbf{Fig. 5:}
\begin{itemize}[leftmargin=1em]
  \item The video features a series of bronze statues in a park, each depicting an adult and a child in various intimate and playful interactions. The statues, set against a backdrop of lush greenery, a wooden bridge, and a clear sky, are placed on pathways and bridges, symbolizing a nurturing relationship. The scenes are tranquil, with no people present, and the soft lighting suggests it's either early morning or late afternoon. The park is well-maintained, with young trees and a serene atmosphere, and the text 'bilibili' appears in one of the frames, indicating a possible association with a media platform.
\end{itemize}

\textbf{Fig. 6:}
\begin{itemize}[leftmargin=1em]
  \item In a dimly lit alleyway, Tom, the mischievous cat, is seen crouching stealthily behind a green trash can. His eyes are wide with anticipation, and his ears are perked up, listening for any sign of Jerry, the clever mouse. The alleyway is littered with various discarded items, creating a sense of clutter and disarray. The lighting casts dramatic shadows, highlighting the tension between the two characters. In the background, the faint sound of footsteps can be heard, adding to the suspenseful atmosphere. The scene captures the classic rivalry between Tom and Jerry, as they engage in their timeless game of cat and mouse.
\end{itemize}

\end{tcolorbox}

% Appendix Box
\begin{tcolorbox}[
    title=Prompt for figures in appendix pages,
    breakable,
    parskip=false,
    % enhanced jigsaw, % <-- 라인 번호 방지에 도움
    sharp corners=south, % <-- 덜 둥근 모서리 (원하는 느낌)
    colback=pastelblue2,
    colframe=pastelblue,
    boxrule=0.5pt,
    arc=1mm,  % 아주 약간 둥글게
    left=6pt, right=6pt, top=6pt, bottom=6pt
]
\small

\textbf{Fig. 9:}
% \ref{fig:apd_crush_hun}
\begin{itemize}[leftmargin=1em]
  \item A rubber boot is placed on the platform. As the hydraulic press moves down, the boot compresses and wrinkles before bursting at the seams.
\end{itemize}

\textbf{Fig. 10:}
% \ref{fig:apd_crush}
\begin{itemize}[leftmargin=1em]
  \item A large, cylindrical object is seen pressing down on a small orange ball, causing it to flatten as if it were under a hydraulic press. The background features a green wall with yellow and red warning signs.
\end{itemize}

\textbf{Fig. 11:}
% \ref{fig:apd_cake_hun}
\begin{itemize}[leftmargin=1em]
  \item A hand wearing a black glove holds a knife, slicing through a Coca-Cola can that has been transformed into a hyper-realistic prop cake. The cake is cut in half, revealing its cake-like interior.
\end{itemize}

\textbf{Fig. 12:}
% \ref{fig:apd_cake_hun2}
\begin{itemize}[leftmargin=1em]
  \item A black leather wallet rests on a wooden surface. A sharp blade cuts into the material, exposing layers of sponge and frosting beneath the realistic edible leather texture.
\end{itemize}

\textbf{Fig. 13:}
% \ref{fig:apd_cake_hun3}
\begin{itemize}[leftmargin=1em]
  \item A bar of soap sits in a soap dish, its pastel color catching the light. A blade smoothly cuts through the bar, revealing layers of lemon cake and frosting beneath the glossy icing.
\end{itemize}

\textbf{Fig. 14:}
% \ref{fig:apd_cake1}
\begin{itemize}[leftmargin=1em]
  \item A stack of pancakes sits on a white plate, drizzled with syrup and butter. A hand with a knife slices through the stack, unveiling that the entire dish is actually a cake, complete with pancake-textured fondant and caramel-flavored layers.
\end{itemize}

\textbf{Fig. 15:}
% \ref{fig:apd_cake3}
\begin{itemize}[leftmargin=1em]
  \item A hand wearing a black glove holds a knife, slicing through a Coca-Cola can that has been transformed into a hyper-realistic prop cake. The cake is cut in half, revealing its cake-like interior.
\end{itemize}

\textbf{Fig. 16:}
% \ref{fig:apd_squish_hun2}
\begin{itemize}[leftmargin=1em]
  \item A steaming cup of coffee sits on a desk. A hand gently presses down on the ceramic mug, which bends and flattens like a stress ball, losing all its rigid form.
\end{itemize}

\textbf{Fig. 17:}
% \ref{fig:apd_squish_hun}
\begin{itemize}[leftmargin=1em]
  \item A soccer ball is placed on a grassy field. A child’s foot kicks it, but instead of bouncing away, it deforms completely and remains squashed on the ground.
\end{itemize}

\textbf{Fig. 18:}
% \ref{fig:apd_squish2}
\begin{itemize}[leftmargin=1em]
  \item A terracotta pot with a visible crack sits centered on a white surface, bathed in sunlight. Two hands enter the frame, positioning themselves around the pot. The hands then begin to press and mold the pot, the clay beginning to rise from the opening. The clay is reshaped and compressed, transforming the pot into a bulbous, amorphous lump. The final shot displays the morphed clay shape standing upright on the white surface.
\end{itemize}

\textbf{Fig. 19:}
% \ref{fig:apd_disney1}
\begin{itemize}[leftmargin=1em]
  \item A black-and-white animated scene unfolds on a semi-rural dock, with a cow standing on wooden planks, holding a piece of paper with 'FOB' written on it. The cow is the central focus, amidst static barrels, crates, and a 'PODUNK LANDING' sign in the background. The atmosphere remains calm and still, with the cow's presence subtly shifting the narrative's tone. A sign of pause or anticipation, the scene is frozen in time, inviting the viewer to ponder the story's next development.
\end{itemize}

\textbf{Fig. 20:}
% \ref{fig:apd_disney}
\begin{itemize}[leftmargin=1em]
  \item A black-and-white animated video showcases a central character with a round body and large ears standing in an indoor setting with a plain background. The character is surrounded by smaller figures, displaying various expressions of interest or curiosity. As the video progresses, subtle changes occur among the figures, suggesting movement and reactions. The scene transitions to focus on a single bird perched on a perch, with its posture and expression changing subtly throughout the frames, showing signs of activity.
\end{itemize}

\textbf{Fig. 21:}
% \ref{fig:apd_tom}
\begin{itemize}[leftmargin=1em]
  \item Tom, the mischievous cat, is crouched in the corner of a dimly lit room, his eyes fixed on Jerry, the clever mouse, who scurries across the wooden floor. The room is adorned with a blue wall and a golden doorknob, adding a touch of elegance to the otherwise mundane setting. Tom's fur is a mix of gray and white, blending seamlessly with the shadows cast by the flickering light. Jerry, on the other hand, is a vibrant mix of brown and white, his nimble movements creating a stark contrast against the stillness of the room. The tension between the two characters is palpable, as they engage in a silent battle of wit and agility. The scene is a classic representation of the timeless rivalry between Tom and Jerry, captured in a moment of suspense and anticipation.
\end{itemize}

\textbf{Fig. 22:}
% \ref{fig:apd_dl3dv}
\begin{itemize}[leftmargin=1em]
  \item The video explores a traditional Chinese museum, starting with a serene interior featuring a glass display case with golden figurines and a bamboo wall with calligraphy. As the video continues, various scenes show the museum's historical artifacts, including a lion statue, a model of an ancient Chinese architectural structure, and a skeleton with animal bones. The exhibit also includes a display case with a human skeleton, a model of an ancient burial site, and a bronze incense burner. The museum's ambiance is tranquil, with soft lighting and a blend of historical and modern elements, culminating in a display of golden figurines and a model of an ancient Chinese architectural complex.
\end{itemize}

\textbf{Fig. 23:}
% \ref{fig:apd_scenes_hun2}
\begin{itemize}[leftmargin=1em]
  \item A person is seen standing in a cluttered room, with a table covered in various items. The scene then shifts to a man holding a gun, seemingly in a threatening stance. The man's presence creates a sense of tension and danger in the room.
\end{itemize}

\textbf{Fig. 24:}
% \ref{fig:apd_scenes_hun}
\begin{itemize}[leftmargin=1em]
  \item In the video, a group of individuals are seen entering a room, each holding a gun. The scene progresses as they move deeper into the room, flattening objects as if they were under a hydraulic press. The room is adorned with framed pictures on the walls, adding to the tension of the scene.
\end{itemize}

\end{tcolorbox}

\newpage
\section{User Study}
\label{app:user_study}
To evaluate the quality of the generated videos, we conducted a human evaluation using 20 samples per criterion. For each sample, participants were shown three videos—generated by Vanilla, REPA*, and CREPA—in a randomized order to ensure fairness. Participants were asked to select the best video for a given criterion, assigning 1 point to their choice. The final score for each model is the total number of times it was selected, with higher scores indicating stronger preference.

\begin{figure}
    \centering
    \includegraphics[width=\textwidth]{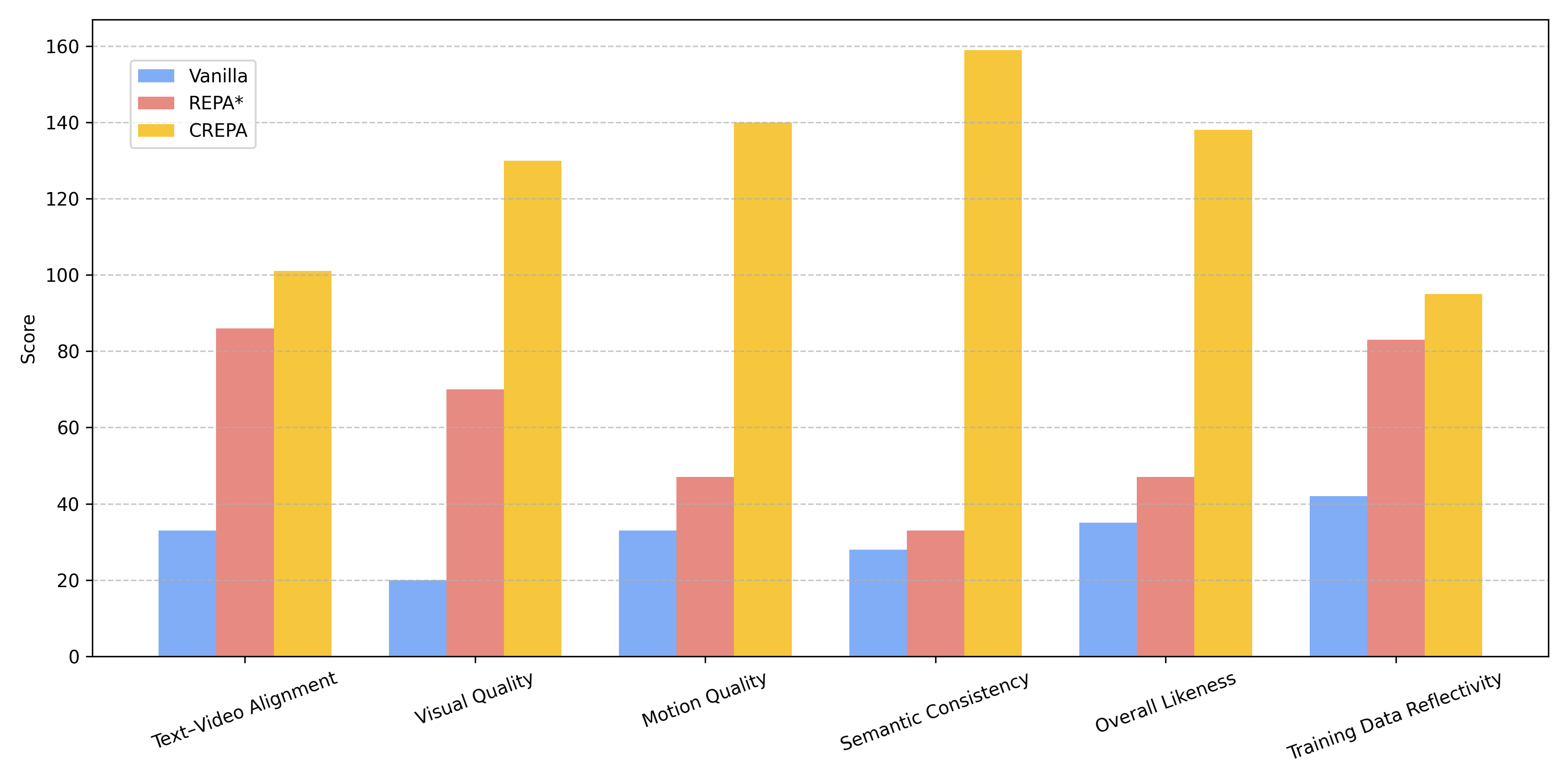}
    \caption{\textbf{Results of user study conducted over Vanilla, REPA*, and CREPA.} CREPA is preferred by human participants over REPA* and Vanilla on all criteria.}
    \label{fig:apd_user_study}
\end{figure}

The evaluation was conducted across six criteria:

(1) Text–Video Alignment, reflecting the frame-level clarity of the video, including sharpness and the absence of artifacts such as tearing or distortion; (2) Visual Quality, reflecting the frame-level clarity of the video, including sharpness and the absence of artifacts such as tearing or distortion;; (3) Motion Quality, evaluating the naturalness and stability of movement across frames, ensuring the motion is smooth and not visually awkward; (4) Overall Likeness, reflecting overall user preference — that is, if a participant had to select a single video for practical use, which one they would choose; (5) Semantic Consistency, measuring whether the video maintains coherent meaning across frames without illogical or abrupt semantic shifts (e.g., object deformation or identity inconsistency); (6) Training Data Reflectivity, assessing how accurately the video reflects the visual and stylistic characteristics of the training distribution; The first five criteria are based on those defined in the EvalCrafter~\cite{evalcrafter} benchmark, with Semantic Consistency adapted from Temporal Consistency to better reflect meaning-level coherence across frames. Training Data Reflectivity was newly introduced to evaluate how well the generated videos capture the visual and stylistic characteristics of the training data. As reported in Fig.~\ref{fig:apd_user_study} , CREPA received the highest scores across all evaluation categories, reflecting a strong overall preference by participants.

\section{Societal Impacts and Safeguards}

Our method, CREPA, offers several positive societal impacts. By reducing data and compute requirements for fine-tuning video diffusion models, it makes high-quality generative tools more accessible to smaller labs, educators, and creators. This can enhance creativity, personalization, and educational applications. However, it also poses risks, including potential misuse for deepfakes, bias amplification from pretrained models, and disruption in creative job markets. These concerns can be mitigated through watermarking and content traceability, ethical usage guidelines, bias audits, and human-in-the-loop deployment to ensure responsible and socially beneficial use.

%% file: main.bbl
\begin{thebibliography}{10}

\bibitem{agarwal2025cosmos}
N.~Agarwal, A.~Ali, M.~Bala, Y.~Balaji, E.~Barker, T.~Cai, P.~Chattopadhyay, Y.~Chen, Y.~Cui, Y.~Ding, et~al.
\newblock Cosmos world foundation model platform for physical ai.
\newblock {\em arXiv preprint arXiv:2501.03575}, 2025.

\bibitem{baker2022vpt}
B.~Baker, I.~Akkaya, P.~Zhokov, J.~Huizinga, J.~Tang, A.~Ecoffet, B.~Houghton, R.~Sampedro, and J.~Clune.
\newblock Video pretraining (vpt): Learning to act by watching unlabeled online videos.
\newblock {\em Advances in Neural Information Processing Systems}, 35:24639--24654, 2022.

\bibitem{bi2025customttt}
X.~Bi, J.~Lu, B.~Liu, X.~Cun, Y.~Zhang, W.~Li, and B.~Xiao.
\newblock Customttt: Motion and appearance customized video generation via test-time training.
\newblock In {\em Proceedings of the AAAI Conference on Artificial Intelligence}, volume~39, pages 1871--1879, 2025.

\bibitem{bigdata-pw_scenes}
bigdata pw.
\newblock Scenes dataset.
\newblock \url{https://huggingface.co/datasets/bigdata-pw/scenes}, 2025.

\bibitem{blattmann2023stable}
A.~Blattmann, T.~Dockhorn, S.~Kulal, D.~Mendelevitch, M.~Kilian, D.~Lorenz, Y.~Levi, Z.~English, V.~Voleti, A.~Letts, et~al.
\newblock Stable video diffusion: Scaling latent video diffusion models to large datasets.
\newblock {\em arXiv preprint arXiv:2311.15127}, 2023.

\bibitem{blattmann2023align}
A.~Blattmann, R.~Rombach, H.~Ling, T.~Dockhorn, S.~W. Kim, S.~Fidler, and K.~Kreis.
\newblock Align your latents: High-resolution video synthesis with latent diffusion models.
\newblock In {\em Proceedings of the IEEE/CVF conference on computer vision and pattern recognition}, pages 22563--22575, 2023.

\bibitem{brown2020language}
T.~Brown, B.~Mann, N.~Ryder, M.~Subbiah, J.~D. Kaplan, P.~Dhariwal, A.~Neelakantan, P.~Shyam, G.~Sastry, A.~Askell, et~al.
\newblock Language models are few-shot learners.
\newblock {\em Advances in neural information processing systems}, 33:1877--1901, 2020.

\bibitem{chen2025jointtuner}
F.~Chen, S.~Zhao, C.~Xu, and L.~Lan.
\newblock Jointtuner: Appearance-motion adaptive joint training for customized video generation.
\newblock {\em arXiv preprint arXiv:2503.23951}, 2025.

\bibitem{deng2009imagenet}
J.~Deng, W.~Dong, R.~Socher, L.-J. Li, K.~Li, and L.~Fei-Fei.
\newblock Imagenet: A large-scale hierarchical image database.
\newblock In {\em 2009 IEEE conference on computer vision and pattern recognition}, pages 248--255. Ieee, 2009.

\bibitem{finetrainers_cakeify_smol}
Finetrainers.
\newblock Cakeify smol dataset.
\newblock \url{https://huggingface.co/datasets/finetrainers/cakeify-smol}, 2024.

\bibitem{finetrainers_crush_smol}
Finetrainers.
\newblock Crush smol dataset.
\newblock \url{https://huggingface.co/datasets/finetrainers/crush-smol}, 2024.

\bibitem{finetrainers_squish_pika}
Finetrainers.
\newblock Squish pika dataset.
\newblock \url{https://huggingface.co/datasets/finetrainers/squish-pika}, 2024.

\bibitem{guo2025long}
Y.~Guo, C.~Yang, Z.~Yang, Z.~Ma, Z.~Lin, Z.~Yang, D.~Lin, and L.~Jiang.
\newblock Long context tuning for video generation.
\newblock {\em arXiv preprint arXiv:2503.10589}, 2025.

\bibitem{gururangan2020don}
S.~Gururangan, A.~Marasovi{\'c}, S.~Swayamdipta, K.~Lo, I.~Beltagy, D.~Downey, and N.~A. Smith.
\newblock Don't stop pretraining: Adapt language models to domains and tasks.
\newblock {\em arXiv preprint arXiv:2004.10964}, 2020.

\bibitem{hecameractrl}
H.~He, Y.~Xu, Y.~Guo, G.~Wetzstein, B.~Dai, H.~Li, and C.~Yang.
\newblock Cameractrl: Enabling camera control for video diffusion models.
\newblock In {\em The Thirteenth International Conference on Learning Representations}.

\bibitem{ho2020denoising}
J.~Ho, A.~Jain, and P.~Abbeel.
\newblock Denoising diffusion probabilistic models.
\newblock {\em Advances in neural information processing systems}, 33:6840--6851, 2020.

\bibitem{hong2024cogvlm2}
W.~Hong, W.~Wang, M.~Ding, W.~Yu, Q.~Lv, Y.~Wang, Y.~Cheng, S.~Huang, J.~Ji, Z.~Xue, et~al.
\newblock Cogvlm2: Visual language models for image and video understanding.
\newblock {\em arXiv preprint arXiv:2408.16500}, 2024.

\bibitem{hu2022lora}
E.~J. Hu, Y.~Shen, P.~Wallis, Z.~Allen-Zhu, Y.~Li, S.~Wang, L.~Wang, W.~Chen, et~al.
\newblock Lora: Low-rank adaptation of large language models.
\newblock {\em ICLR}, 1(2):3, 2022.

\bibitem{hu2023efficiently}
X.~Hu, X.~Xu, and Y.~Shi.
\newblock How to efficiently adapt large segmentation model (sam) to medical images.
\newblock {\em arXiv preprint arXiv:2306.13731}, 2023.

\bibitem{huang2024vbench}
Z.~Huang, Y.~He, J.~Yu, F.~Zhang, C.~Si, Y.~Jiang, Y.~Zhang, T.~Wu, Q.~Jin, N.~Chanpaisit, et~al.
\newblock Vbench: Comprehensive benchmark suite for video generative models.
\newblock In {\em Proceedings of the IEEE/CVF Conference on Computer Vision and Pattern Recognition}, pages 21807--21818, 2024.

\bibitem{huh2024platonic}
M.~Huh, B.~Cheung, T.~Wang, and P.~Isola.
\newblock The platonic representation hypothesis.
\newblock {\em CoRR}, 2024.

\bibitem{kerbl20233d}
B.~Kerbl, G.~Kopanas, T.~Leimk{\"u}hler, and G.~Drettakis.
\newblock 3d gaussian splatting for real-time radiance field rendering.
\newblock {\em ACM Trans. Graph.}, 42(4):139--1, 2023.

\bibitem{kong2024hunyuanvideo}
W.~Kong, Q.~Tian, Z.~Zhang, R.~Min, Z.~Dai, J.~Zhou, J.~Xiong, X.~Li, B.~Wu, J.~Zhang, et~al.
\newblock Hunyuanvideo: A systematic framework for large video generative models.
\newblock {\em arXiv preprint arXiv:2412.03603}, 2024.

\bibitem{kornblith2019similarity}
S.~Kornblith, M.~Norouzi, H.~Lee, and G.~Hinton.
\newblock Similarity of neural network representations revisited.
\newblock In {\em International conference on machine learning}, pages 3519--3529. PMLR, 2019.

\bibitem{liang2024wonderland}
H.~Liang, J.~Cao, V.~Goel, G.~Qian, S.~Korolev, D.~Terzopoulos, K.~N. Plataniotis, S.~Tulyakov, and J.~Ren.
\newblock Wonderland: Navigating 3d scenes from a single image.
\newblock {\em arXiv preprint arXiv:2412.12091}, 2024.

\bibitem{ling2024dl3dv}
L.~Ling, Y.~Sheng, Z.~Tu, W.~Zhao, C.~Xin, K.~Wan, L.~Yu, Q.~Guo, Z.~Yu, Y.~Lu, et~al.
\newblock Dl3dv-10k: A large-scale scene dataset for deep learning-based 3d vision.
\newblock In {\em Proceedings of the IEEE/CVF Conference on Computer Vision and Pattern Recognition}, pages 22160--22169, 2024.

\bibitem{liu2024improved}
H.~Liu, C.~Li, Y.~Li, and Y.~J. Lee.
\newblock Improved baselines with visual instruction tuning.
\newblock In {\em Proceedings of the IEEE/CVF Conference on Computer Vision and Pattern Recognition}, pages 26296--26306, 2024.

\bibitem{liu2023visual}
H.~Liu, C.~Li, Q.~Wu, and Y.~J. Lee.
\newblock Visual instruction tuning.
\newblock {\em Advances in neural information processing systems}, 36:34892--34916, 2023.

\bibitem{evalcrafter}
Y.~Liu, X.~Cun, X.~Liu, X.~Wang, Y.~Zhang, H.~Chen, Y.~Liu, T.~Zeng, R.~Chan, and Y.~Shan.
\newblock Evalcrafter: Benchmarking and evaluating large video generation models.
\newblock In {\em Proceedings of the IEEE/CVF Conference on Computer Vision and Pattern Recognition}, pages 22139--22149, 2024.

\bibitem{lu2024lora3d}
Z.~Lu, H.~Yang, D.~Xu, B.~Li, B.~Ivanovic, M.~Pavone, and Y.~Wang.
\newblock Lora3d: Low-rank self-calibration of 3d geometric foundation models.
\newblock {\em arXiv preprint arXiv:2412.07746}, 2024.

\bibitem{ma2024sit}
N.~Ma, M.~Goldstein, M.~S. Albergo, N.~M. Boffi, E.~Vanden-Eijnden, and S.~Xie.
\newblock Sit: Exploring flow and diffusion-based generative models with scalable interpolant transformers.
\newblock In {\em European Conference on Computer Vision}, pages 23--40. Springer, 2024.

\bibitem{mahendran2015understanding}
A.~Mahendran and A.~Vedaldi.
\newblock Understanding deep image representations by inverting them.
\newblock In {\em Proceedings of the IEEE conference on computer vision and pattern recognition}, pages 5188--5196, 2015.

\bibitem{oquabdinov2}
M.~Oquab, T.~Darcet, T.~Moutakanni, H.~V. Vo, M.~Szafraniec, V.~Khalidov, P.~Fernandez, D.~HAZIZA, F.~Massa, A.~El-Nouby, et~al.
\newblock Dinov2: Learning robust visual features without supervision.
\newblock {\em Transactions on Machine Learning Research}.

\bibitem{peebles2023scalable}
W.~Peebles and S.~Xie.
\newblock Scalable diffusion models with transformers.
\newblock In {\em Proceedings of the IEEE/CVF international conference on computer vision}, pages 4195--4205, 2023.

\bibitem{rudin1992nonlinear}
L.~I. Rudin, S.~Osher, and E.~Fatemi.
\newblock Nonlinear total variation based noise removal algorithms.
\newblock {\em Physica D: nonlinear phenomena}, 60(1-4):259--268, 1992.

\bibitem{ruiz2023dreambooth}
N.~Ruiz, Y.~Li, V.~Jampani, Y.~Pritch, M.~Rubinstein, and K.~Aberman.
\newblock Dreambooth: Fine tuning text-to-image diffusion models for subject-driven generation.
\newblock In {\em Proceedings of the IEEE/CVF conference on computer vision and pattern recognition}, pages 22500--22510, 2023.

\bibitem{schoenberger2016sfm}
J.~L. Sch\"{o}nberger and J.-M. Frahm.
\newblock Structure-from-motion revisited.
\newblock In {\em Conference on Computer Vision and Pattern Recognition (CVPR)}, 2016.

\bibitem{shah2024ziplora}
V.~Shah, N.~Ruiz, F.~Cole, E.~Lu, S.~Lazebnik, Y.~Li, and V.~Jampani.
\newblock Ziplora: Any subject in any style by effectively merging loras.
\newblock In {\em European Conference on Computer Vision}, pages 422--438. Springer, 2024.

\bibitem{sohl2015deep}
J.~Sohl-Dickstein, E.~Weiss, N.~Maheswaranathan, and S.~Ganguli.
\newblock Deep unsupervised learning using nonequilibrium thermodynamics.
\newblock In {\em International conference on machine learning}, pages 2256--2265. PMLR, 2015.

\bibitem{song2020denoising}
J.~Song, C.~Meng, and S.~Ermon.
\newblock Denoising diffusion implicit models.
\newblock {\em arXiv preprint arXiv:2010.02502}, 2020.

\bibitem{song2020score}
Y.~Song, J.~Sohl-Dickstein, D.~P. Kingma, A.~Kumar, S.~Ermon, and B.~Poole.
\newblock Score-based generative modeling through stochastic differential equations.
\newblock {\em arXiv preprint arXiv:2011.13456}, 2020.

\bibitem{wang2025wan}
A.~Wang, B.~Ai, B.~Wen, C.~Mao, C.-W. Xie, D.~Chen, F.~Yu, H.~Zhao, J.~Yang, J.~Zeng, et~al.
\newblock Wan: Open and advanced large-scale video generative models.
\newblock {\em arXiv preprint arXiv:2503.20314}, 2025.

\bibitem{wang2024dust3r}
S.~Wang, V.~Leroy, Y.~Cabon, B.~Chidlovskii, and J.~Revaud.
\newblock Dust3r: Geometric 3d vision made easy.
\newblock In {\em Proceedings of the IEEE/CVF Conference on Computer Vision and Pattern Recognition}, pages 20697--20709, 2024.

\bibitem{wang2023stylediffusion}
Z.~Wang, L.~Zhao, and W.~Xing.
\newblock Stylediffusion: Controllable disentangled style transfer via diffusion models.
\newblock In {\em Proceedings of the IEEE/CVF International Conference on Computer Vision}, pages 7677--7689, 2023.

\bibitem{wildheart2024disney}
Wild-Heart.
\newblock Disney video generation dataset.
\newblock \url{https://huggingface.co/datasets/Wild-Heart/Disney-VideoGeneration-Dataset}, 2024.

\bibitem{wildheart2024tomjerry}
Wild-Heart.
\newblock Tom and jerry video generation dataset.
\newblock \url{https://huggingface.co/datasets/Wild-Heart/Tom-and-Jerry-VideoGeneration-Dataset}, 2024.

\bibitem{xiang2023denoising}
W.~Xiang, H.~Yang, D.~Huang, and Y.~Wang.
\newblock Denoising diffusion autoencoders are unified self-supervised learners.
\newblock In {\em Proceedings of the IEEE/CVF International Conference on Computer Vision}, pages 15802--15812, 2023.

\bibitem{yang2024cogvideox}
Z.~Yang, J.~Teng, W.~Zheng, M.~Ding, S.~Huang, J.~Xu, Y.~Yang, W.~Hong, X.~Zhang, G.~Feng, et~al.
\newblock Cogvideox: Text-to-video diffusion models with an expert transformer.
\newblock {\em CoRR}, 2024.

\bibitem{yeh2022total}
R.~A. Yeh, C.-Y. Chen, and A.~G. Schwing.
\newblock Total variation optimization layers for computer vision.
\newblock In {\em Proceedings of the IEEE/CVF Conference on Computer Vision and Pattern Recognition (CVPR)}, pages 12345--12354, 2022.

\bibitem{yu2024representation}
S.~Yu, S.~Kwak, H.~Jang, J.~Jeong, J.~Huang, J.~Shin, and S.~Xie.
\newblock Representation alignment for generation: Training diffusion transformers is easier than you think.
\newblock {\em arXiv preprint arXiv:2410.06940}, 2024.

\bibitem{yue2024improving}
Y.~Yue, A.~Das, F.~Engelmann, S.~Tang, and J.~E. Lenssen.
\newblock Improving 2d feature representations by 3d-aware fine-tuning.
\newblock In {\em European Conference on Computer Vision}, pages 57--74. Springer, 2024.

\bibitem{zheng2024open}
Z.~Zheng, X.~Peng, T.~Yang, C.~Shen, S.~Li, H.~Liu, Y.~Zhou, T.~Li, and Y.~You.
\newblock Open-sora: Democratizing efficient video production for all.
\newblock {\em arXiv preprint arXiv:2412.20404}, 2024.

\end{thebibliography}
